\documentclass{article}

\PassOptionsToPackage{numbers, square, compress}{natbib}
\usepackage[preprint]{neurips_2026}
\usepackage[utf8]{inputenc}
\usepackage[T1]{fontenc}
\usepackage{hyperref}
\usepackage{url}
\usepackage{booktabs}
\usepackage{amsfonts}
\usepackage{nicefrac}
\usepackage{microtype}
\usepackage{xcolor}
\usepackage{multirow}
\usepackage{graphicx}
\usepackage{adjustbox}
\usepackage{wrapfig}
\usepackage{array}
\usepackage{CJKutf8}
\usepackage{caption}
\usepackage{amsmath}
\usepackage{listings}
\usepackage{enumitem}

\graphicspath{{figs/}}

\title{PRISM: Prior Rectification and Uncertainty-Aware Structure Modeling for Diffusion-Based \\ Text Image Super-Resolution}

\author{%
  Zihang Xu$^{1}$\thanks{Equal contribution}, \enspace 
  Xiaoyang Liu$^{1}$\footnotemark[1], \enspace
  Zheng Chen$^{1}$, \enspace
  \textbf{Yulun Zhang$^{1}$}\thanks{Corresponding author: Yulun Zhang, yulun100@gmail.com}, \enspace
  \textbf{Xiaokang Yang$^{1}$} \enspace \\
  \textsuperscript{1}Shanghai Jiao Tong University \enspace
}

\begin{document}

\maketitle

\vspace{-4mm}%
\begin{abstract}
Text image super-resolution (Text-SR) requires more than visually plausible detail synthesis: slight errors in stroke topology may alter character identity and break readability. Existing methods improve text fidelity with stronger recognition-based or generative priors, yet they still face two unresolved challenges under severe degradation: the text condition extracted from low-quality inputs can itself be unreliable, and a plausible global prior does not fully determine fine-grained stroke boundaries. We present \textbf{PRISM}, a single-step diffusion-based Text-SR framework that addresses these two challenges through Flow-Matching Prior Rectification (\textbf{FMPR}) and a Structure-guided Uncertainty-aware Residual Encoder (\textbf{SURE}). FMPR constructs a privileged training-time prior from paired low-quality/high-quality latents and learns a flow matching that transports degraded embeddings toward this restoration-oriented prior space, yielding more accurate and reliable global text guidance. SURE further predicts uncertainty-aware structural residuals to selectively absorb reliable local boundary evidence while suppressing ambiguous stroke cues. Together, these components enable explicit global prior rectification and local structure refinement within a single diffusion restoration pass. Experiments on both synthetic and real-world benchmarks show that PRISM achieves state-of-the-art performance with millisecond-level inference. Our dataset and code will be available at~\url{https://github.com/faithxuz/PRISM}.
\end{abstract}

\vspace{-4mm}
\section{Introduction}
\vspace{-3mm}
Text image super-resolution (Text-SR) aims to restore high-resolution text images from degraded low-resolution inputs. Unlike generic image super-resolution~\citep{wu2024osediff,dong2025tsdsr,li2025fluxsr}, text is both visual and symbolic. A small artifact in a natural texture may only affect perceptual quality, whereas a broken stroke, merged component, or distorted enclosure can change the identity of a character. This sensitivity is especially severe for densely structured scripts such as Chinese~\citep{li2023marconet}, where subtle stroke layouts often distinguish different characters. An effective Text-SR system must therefore recover not only visually plausible details, but also semantically faithful glyph structures with sub-character precision.

Existing Text-SR methods address this structure-sensitive problem by introducing stronger text-specific guidance. Early methods~\citep{wang2020tsrn,chen2021tbsrn,ma2022tatt} improve readability with recognition supervision, sequential modeling, and layout-aware reasoning. These cues help the model reason about text, but can become unreliable when severe degradation removes stroke evidence needed for character discrimination. Later methods~\citep{li2023marconet,yuan2025stylesrn} introduce richer text-specific priors, such as generative character-structure priors and text style embeddings, to handle complex glyphs and appearance variation. Recent diffusion-based Text-SR and text-aware restoration methods~\citep{zhang2024difftsr,hu2025tadisr,min2026terediff,he2026textsdiff} further exploit generative priors, text diffusion, segmentation, or text-spotting guidance to improve perceptual realism and text fidelity. While these developments highlight the importance of text-aware guidance, its reliability under severe degradation and its effective translation into local stroke geometry remain insufficiently addressed.

\begin{wrapfigure}[11]{r}{0.56\linewidth}
\vspace{0em}
\centering

\setlength{\tabcolsep}{0pt}
\renewcommand{\arraystretch}{0}

\begin{adjustbox}{max width=\linewidth}
\begin{tabular}{@{}r@{\hspace{1mm}}c@{\hspace{0.8mm}}c@{\hspace{0.8mm}}c@{}}

\raisebox{-.5\height}{\makebox[0.1\linewidth][r]{\scriptsize GT}} &
\raisebox{-.5\height}{\includegraphics[height=0.086\linewidth,keepaspectratio]{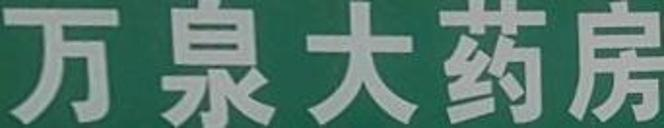}} &
\raisebox{-.5\height}{\includegraphics[height=0.086\linewidth,keepaspectratio]{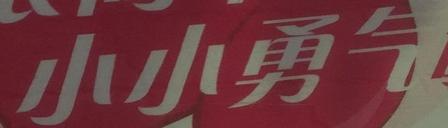}} &
\raisebox{-.5\height}{\includegraphics[height=0.086\linewidth,keepaspectratio]{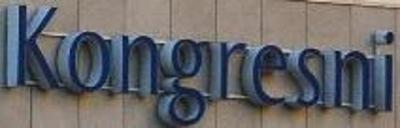}}
\vspace{1mm}
\\[1.5pt]

\raisebox{-.5\height}{\makebox[0.13\linewidth][r]{\scriptsize LR}} &
\raisebox{-.5\height}{\includegraphics[height=0.086\linewidth,keepaspectratio]{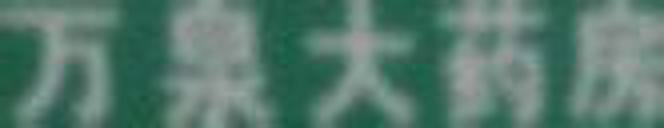}} &
\raisebox{-.5\height}{\includegraphics[height=0.086\linewidth,keepaspectratio]{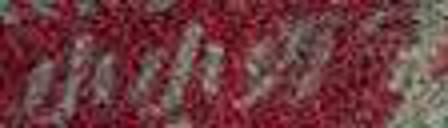}} &
\raisebox{-.5\height}{\includegraphics[height=0.086\linewidth,keepaspectratio]{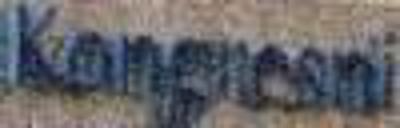}}
\vspace{1mm}
\\[1.5pt]

\raisebox{-.5\height}{\makebox[0.13\linewidth][r]{\scriptsize MARCONet}} &
\raisebox{-.5\height}{\includegraphics[height=0.086\linewidth,keepaspectratio]{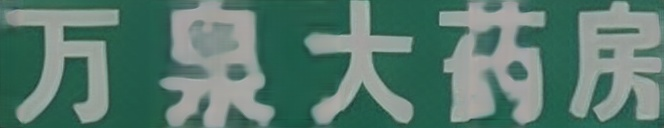}} &
\raisebox{-.5\height}{\includegraphics[height=0.086\linewidth,keepaspectratio]{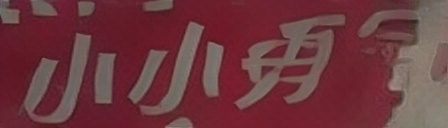}} &
\raisebox{-.5\height}{\includegraphics[height=0.086\linewidth,keepaspectratio]{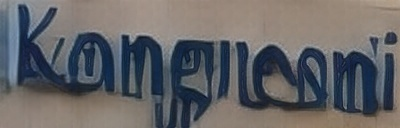}}
\vspace{1mm}
\\[1.5pt]

\raisebox{-.5\height}{\makebox[0.13\linewidth][r]{\scriptsize DiffTSR}} &
\raisebox{-.5\height}{\includegraphics[height=0.086\linewidth,keepaspectratio]{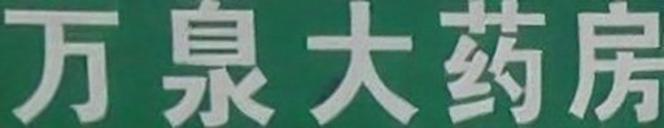}} &
\raisebox{-.5\height}{\includegraphics[height=0.086\linewidth,keepaspectratio]{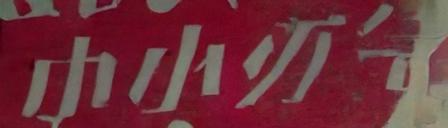}} &
\raisebox{-.5\height}{\includegraphics[height=0.086\linewidth,keepaspectratio]{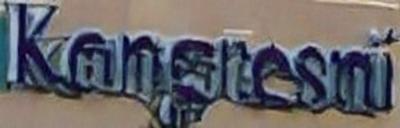}}
\vspace{1mm}
\\[1.5pt]

\raisebox{-.5\height}{\makebox[0.13\linewidth][r]{\scriptsize TeReDiff}} &
\raisebox{-.5\height}{\includegraphics[height=0.086\linewidth,keepaspectratio]{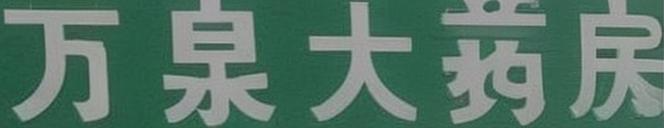}} &
\raisebox{-.5\height}{\includegraphics[height=0.086\linewidth,keepaspectratio]{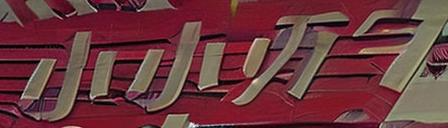}} &
\raisebox{-.5\height}{\includegraphics[height=0.086\linewidth,keepaspectratio]{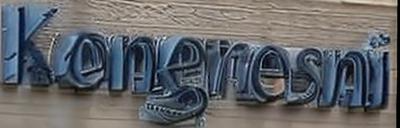}}
\vspace{1mm}
\\[1.5pt]

\raisebox{-.5\height}{\makebox[0.13\linewidth][r]{\scriptsize PRISM}} &
\raisebox{-.5\height}{\includegraphics[height=0.086\linewidth,keepaspectratio]{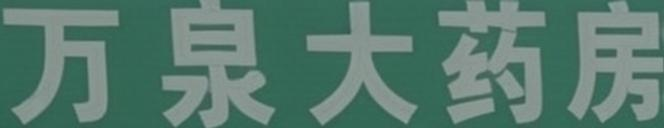}} &
\raisebox{-.5\height}{\includegraphics[height=0.086\linewidth,keepaspectratio]{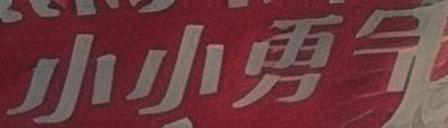}} &
\raisebox{-.5\height}{\includegraphics[height=0.086\linewidth,keepaspectratio]{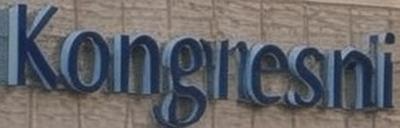}}

\end{tabular}
\end{adjustbox}

\vspace{-0.6em}
\caption{Entangled objectives cause structural errors.}
\label{fig:motivation}
\vspace{-1mm}
\end{wrapfigure}

Instead of debating whether to incorporate text-aware cues, the current bottleneck lies in how to obtain them reliably under severe degradation. In recent diffusion-based methods, text conditions are typically derived directly from the degraded input. When strokes are heavily corrupted, these inferred conditions are inherently unreliable. Because condition estimation and image reconstruction are entangled under a shared objective, the model cannot distinguish between correcting stroke geometry and compensating for an erroneous high-level condition, often yielding sharp but semantically incorrect outputs (Fig.~\ref{fig:motivation}). Moreover, even if a plausible global semantic condition is obtained, it cannot fully determine pixel-aligned local structures such as stroke closures and intersections. Directly relying on edge cues from the degraded image to fill this gap is equally risky, as the visible edges are often missing or misleading. These coupled challenges suggest a need for an explicit decomposition: we could first recover a stable text-aware latent condition from the degraded input, and subsequently refine uncertain local stroke geometry in image space under that guidance.

We propose \textbf{PRISM}, a single-step Text-SR framework based on pre-trained Diffusion Models (DMs), with \textbf{P}rior \textbf{R}ectification and uncerta\textbf{I}nty-aware \textbf{S}tructure \textbf{M}odeling. PRISM explicitly decomposes restoration into global prior rectification and local structure refinement. Its first component, \textbf{FMPR} (\textbf{F}low-\textbf{M}atching \textbf{P}rior \textbf{R}ectification), constructs a privileged training-time prior from paired LQ/HQ latents and learns a flow matching that transports the LQ embedding distribution toward this privileged prior space. Unlike conventional diffusion-style prior extraction that starts from pure noise or treats the inferred prior as a static side condition, FMPR directly models the velocity field from degraded embeddings to restoration-oriented text tokens, producing more accurate and reliable global guidance. 

The second component, \textbf{SURE} (\textbf{S}tructure-guided \textbf{U}ncertainty-aware \textbf{R}esidual \textbf{E}ncoder), injects residual controls to refine local stroke geometry. SURE is a structure-aware encoder branch that predicts both the mean and uncertainty of structural features, allowing the model to selectively absorb reliable boundaries while suppressing ambiguous ones, instead of treating LQ edges as deterministic truth. This uncertainty-aware design is particularly important for Text-SR, where an overconfident wrong edge can be more harmful than a missing edge. To the best of our knowledge, this is the first uncertainty-aware boundary control formulation tailored to text-specific structural refinement.

PRISM keeps the efficiency advantage of one-step restoration while substantially improving the quality of text-aware guidance and structure recovery. The FMPR flow transport is performed in a compact embedding space, and the final image restoration still uses a single diffusion backbone call, making the overall system significantly faster than iterative diffusion-based Text-SR while preserving superior generative quality. Experiments on both synthetic and real-world benchmarks show that PRISM achieves state-of-the-art overall performance with millisecond-level inference.

Our contributions are summarized as follows:
\vspace{-2mm}
\begin{itemize}[left=1em, itemsep=-1pt]
    \item We revisit Text-SR from the perspective of \emph{prior reliability} and \emph{structural uncertainty}, and propose \textbf{PRISM}, a Text-SR model with single-step diffusion inference.
    \item We propose \textbf{FMPR}, a flow-matching prior rectification module that learns to transport LQ text embeddings toward a privileged HQ-aware prior space and injects the recovered tokens into the main backbone for efficient restoration.
    \item We propose \textbf{SURE}, an uncertainty-aware structure guidance module that predicts stochastic edge features and adaptively gates boundary information through uncertainty learning, yielding more robust local structure control under severe degradation.
    \item Extensive experiments on both synthetic and real-world benchmarks show that PRISM achieves state-of-the-art performance at the millisecond level.
\end{itemize}

\vspace{-2mm}
\section{Related Works}
\label{sec:related_works}
\vspace{-2mm}
\noindent \textbf{Real-World Image Super-Resolution.}
Real-world image super-resolution (Real-SR) aims to restore high-quality images from low-resolution inputs with complex and unknown degradations. Early methods mainly improve robustness through degradation modeling and discriminative reconstruction, such as BSRGAN~\citep{zhang2021bsrgan} and Real-ESRGAN~\citep{wang2021realesrgan}. With the development of generative models~\citep{ldm}, recent methods exploit diffusion priors to recover realistic details under severe degradation~\citep{wang2024stablesr,lin2024diffbir, liu2025osdd, liu2025fidediff, wu2024seesr, yu2024supir}. For example, DiffBIR~\citep{lin2024diffbir} decomposes blind restoration into degradation removal and diffusion-based detail regeneration, while SUPIR~\citep{yu2024supir} scales generative restoration with large diffusion priors and high-quality data. Since iterative diffusion sampling is expensive, efficient Real-SR methods further compress or reformulate diffusion restoration into few-step or one-step inference~\citep{wu2024osediff,dong2025tsdsr,li2025fluxsr,wang2024sinsr,yue2025invsr,lin2025hypir}. OSEDiff~\citep{wu2024osediff}, for instance, performs one-step Real-SR by directly starting from the low-quality image. Stronger generative backbones, including SDXL~\citep{podell2023sdxl}, DiT~\citep{peebles2023dit}, SD3~\citep{esser2024sd3}, and FLUX~\citep{flux2024}, have also been studied or adapted for restoration~\citep{duan2025dit4sr,li2025fluxsr}. However, these methods mainly target generic natural image restoration and lack dedicated modeling for character identity and stroke structure.

\noindent \textbf{Text Image Super-Resolution.}
Text image super-resolution (Text-SR) focuses on restoring readable text crops or text-line images from degraded inputs. Different from generic SR, Text-SR requires the restored image to preserve character identity as well as visual quality. Early methods address this problem by introducing recognition guidance, sequential reasoning, layout modeling, and text-prior attention~\citep{wang2020tsrn,chen2021tbsrn,ma2022tatt,ma2023tpgsr,zhao2023stirer}. TSRN~\citep{wang2020tsrn} frames Text-SR as a recognition-oriented restoration problem, while TBSRN~\citep{chen2021tbsrn} and TATT~\citep{ma2022tatt} further exploit text layouts, character details, and deformation-aware attention. Later studies move from high-level recognition cues toward more explicit text structure modeling~\citep{li2023marconet,yuan2025stylesrn,guo2023lemma,zhu2023tsan,zhu2023dpmn,zhao2024pean,wei2025glyphsr,li2025marconetplusplus}. These works shift the focus from recognizing text to preserving how characters are spatially organized and visually presented. MARCONet~\citep{li2023marconet} learns a generative structure prior for blind text restoration, while StyleSRN~\citep{yuan2025stylesrn} complements text priors with style embeddings to better preserve appearance details. More recently, diffusion-driven Text-SR methods have explored generative restoration under text-specific conditions~\citep{singh2024dcdm,zhang2024difftsr}. DiffTSR~\citep{zhang2024difftsr} couples image and text diffusion, demonstrating the potential of diffusion priors for severely degraded Text-SR.

A closely related direction studies text-aware restoration in broader real-world or full-image settings~\citep{hu2025tadisr,min2026terediff,he2026textsdiff}. These methods usually build upon general restoration frameworks and introduce text awareness through text-region perception, segmentation, text spotting, or text-aware conditioning. TADiSR~\citep{hu2025tadisr} integrates text-aware attention and joint segmentation decoders for real-world image SR, while TeReDiff~\citep{min2026terediff} couples diffusion restoration with a text-spotting module. Although these works operate on full images, their text-related component is closely connected to crop-level Text-SR: full-image text-aware restoration still requires reliable restoration of local text regions, while crop-level Text-SR isolates this text-centric subproblem and enables more focused modeling of character fidelity and stroke structures. Thus, the two settings are mutually convertible and complementary. Following this rationale, we adopt the crop-level setting and focus on text-line super-resolution. By isolating the problem at the crop level, we are able to design highly dedicated modules for reliable text prior recovery and uncertainty-aware stroke refinement. Furthermore, our method can be seamlessly integrated into full-image restoration pipelines as a robust, dedicated text-enhancing module.

\vspace{-2mm}
\section{Methodology}
\label{sec:methodology}
\vspace{-2mm}
 
\subsection{Overall Structure}
\label{sec:overall_structure}
\vspace{-2mm}
\begin{figure}[t]
    \centering
    \includegraphics[width=\linewidth]{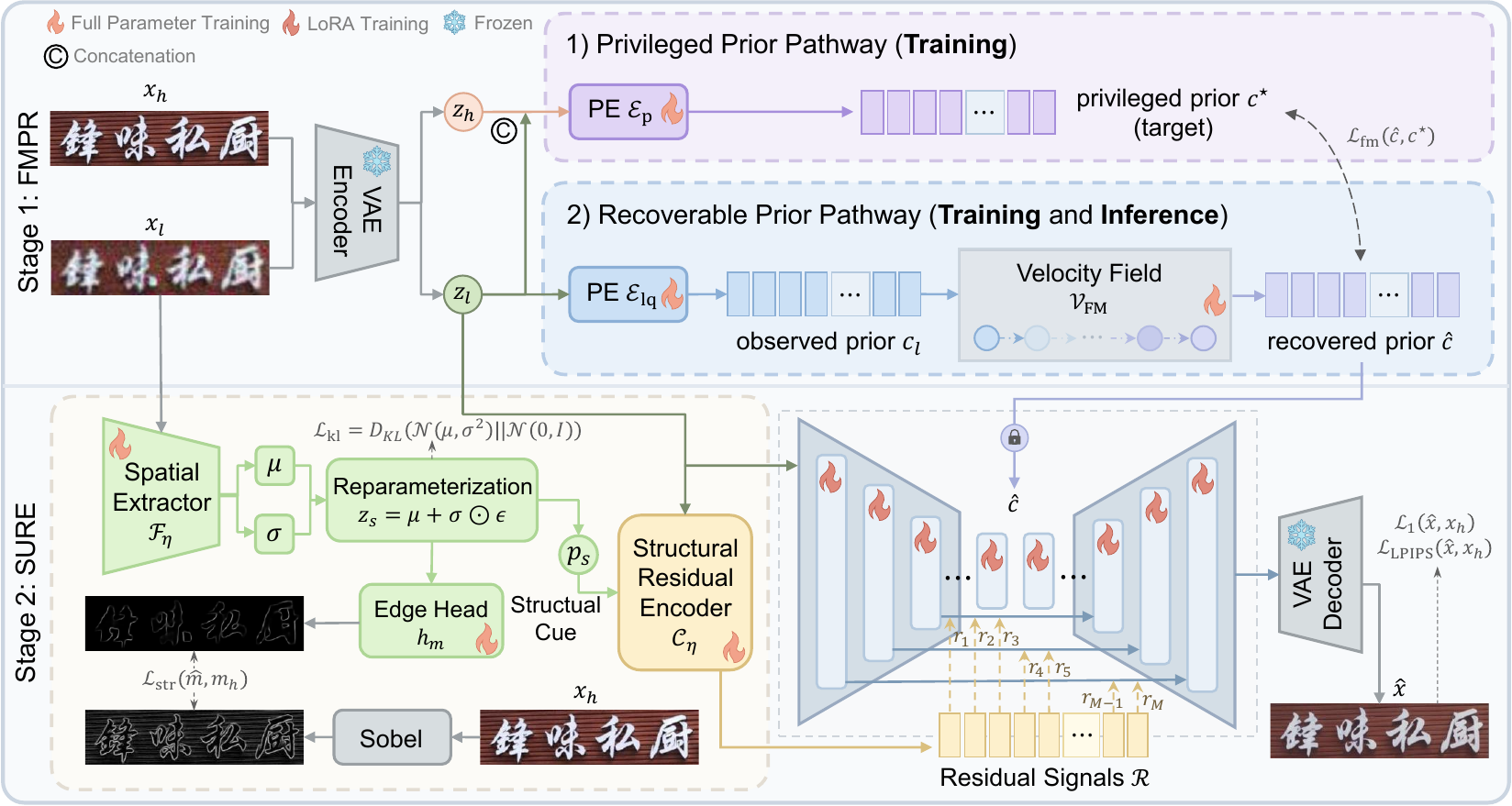}
    \vspace{-4mm}
    \caption{Overall structure of our PRISM.}
    \label{fig:pipeline}
    \vspace{-6mm}
\end{figure}

The overall structure of our \textbf{PRISM} is illustrated in Fig.~\ref{fig:pipeline}. Built upon a pre-trained latent diffusion model~\citep{ldm}, our method follows a progressive restoration paradigm for Text-SR. Severe text degradation introduces two coupled challenges: the text-aware condition inferred from the degraded input may be unreliable, while fine-grained stroke topology and boundary placement may remain ambiguous even with a plausible prior. To address these, we first learn a recoverable text prior and then refine spatially unstable structures under the recovered prior.

Given a degraded text image $x_l$, the frozen VAE encoder maps it to a latent representation $z_l=\mathcal{E}_{\mathrm{vae}}(x_l)$. The prior recovery branch \textbf{FMPR} predicts a text-aware embedding $\hat{c}$ from $z_l$, where $\hat{c}$ is learned to approximate a privileged prior space constructed from paired training data. In parallel, the structure control branch \textbf{SURE} extracts uncertainty-aware spatial cues from $x_l$ and predicts multi-level residual controls $\mathcal{R}=\{r_i\}_{i=1}^{M}$. Following~\citep{wu2024osediff, li2024distillation}, the single-step restoration is computed as $\hat{z}_h = \frac{z_l - \sqrt{1 - \bar{\alpha}_t} \hat{\epsilon}}{\sqrt{\bar{\alpha}_t}}$~\citep{ddpm, ldm}, where $z_l$ is the degraded latent at a fixed timestep $t$, $\bar{\alpha}_t$ is the noise schedule coefficient, and $\hat{\epsilon}$ is the predicted noise. For brevity, we denote the overall process as:
\begin{equation}
    \hat{z}_h
    =
    \mathcal{U}_{\bar{\theta}}
    \left(
        z_l, \hat{c}; \mathcal{R}
    \right),
    \qquad
    \hat{x}
    =
    \mathcal{D}_{\mathrm{vae}}(\hat{z}_h),
    \label{eq:overall_pipeline}
\end{equation}
where $\mathcal{U}_{\bar{\theta}}$ denotes the diffusion backbone used in the final stage, and $\mathcal{D}_{\mathrm{vae}}$ is the VAE decoder. For clarity, we use $\theta_{\mathrm{p}}$, $\theta_{\mathrm{r}}$, and $\bar{\theta}$ to denote the restoration backbone after privileged-prior construction, after recoverable-prior learning, and after training for structure control, respectively.

During training, we first construct a privileged conditional prior from paired LQ/HQ latents and learn to recover it from the degraded input alone. After the recoverable prior pathway is trained, we freeze both the prior pathway and the restoration backbone and optimize the structure control branch. During inference, the model only requires the degraded input $x_l$: the prior branch produces $\hat{c}$, the structure branch produces $\mathcal{R}$, and the restoration backbone generates the final output. 

\vspace{-2mm}
\subsection{FMPR: Flow-Matching Prior Rectification}
\label{sec:recoverable_prior}
\vspace{-2mm}
A reliable text-aware condition is crucial for Text-SR but difficult to obtain under severe degradation. Direct extraction from degraded images often yields unreliable priors that misguide restoration. Thus, our goal is not merely to apply a text prior, but to learn one that is informative during training and recoverable from degraded observations at test time.

Our solution, FMPR, decouples prior construction from prior recovery. During training, paired high-quality and low-quality data allow us to construct a privileged conditional prior that defines a target prior space. At inference, where only degraded inputs are available, we learn an LQ-only recovery path to map observations toward this privileged space. This follows the spirit of learning with privileged information~\citep{vapnik2015learning,lee2020pisr,xia2023diffir}: extra information available only during training defines a more reliable learning target, while the inference model remains dependent solely on observed inputs.

\paragraph{Privileged Conditional Prior.}
Given a paired training sample $(x_l,x_h)$, we encode both images into the latent space as $z_l=\mathcal{E}_{\mathrm{vae}}(x_l)$ and $z_h=\mathcal{E}_{\mathrm{vae}}(x_h)$ with the frozen VAE encoder. A prior encoder (PE) $\mathcal{E}_{\mathrm{p}}$ takes the concatenated LQ-HQ latents and produces a privileged conditional prior. The privileged-prior construction is formulated as
\begin{equation}
    c^{\star}
    =
    \mathcal{E}_{\mathrm{p}}([z_l;z_h]),
    \qquad
    c^{\star}\in\mathbb{R}^{N\times D},
    \label{eq:privileged_prior_compact}
\end{equation}
where $[\cdot;\cdot]$ denotes channel-wise concatenation and $N$, $D$ are the token number and channel dimension. Since $c^{\star}$ sees both degraded evidence and target latent structure, it provides a cleaner conditional signal than an LQ-only prior. We use it as the text embedding to warm up the one-step backbone, where $c^{\star}$ serves as the key (K) and value (V) for the UNet cross-attention layers:
\begin{equation}
    \hat{z}^{\star}_h
    =
    \mathcal{U}_{\theta_{\mathrm{p}}}(z_l,c^{\star}),
    \qquad
    \hat{x}^{\star}
    =
    \mathcal{D}_{\mathrm{vae}}(\hat{z}^{\star}_h),
    \qquad
    \mathcal{L}_{\mathrm{priv}}
    =
    \|\hat{x}^{\star}-x_h\|_1
    +
    \lambda_{\mathrm{lpips}}
    \mathcal{L}_{\mathrm{LPIPS}}(\hat{x}^{\star},x_h).
    \label{eq:privileged_loss}
\end{equation}
Importantly, $c^{\star}$ is only available during training; it defines the target prior distribution rather than a test-time condition. Its role is to define a privileged prior space that specifies what an informative text-aware condition should look like for restoration.

\paragraph{Recoverable Prior Learning.}
After the privileged prior space is established, the remaining problem is how to approximate it without access to $x_h$. We first map the degraded latent to an observed prior $c_l=\mathcal{E}_{\mathrm{lq}}(z_l)$ using an LQ-only PE $\mathcal{E}_{\mathrm{lq}}$ with the same structure as $\mathcal{E}_{\mathrm{p}}$.
\begin{wrapfigure}{r}{0.45\linewidth}
\centering
\includegraphics[width=\linewidth]{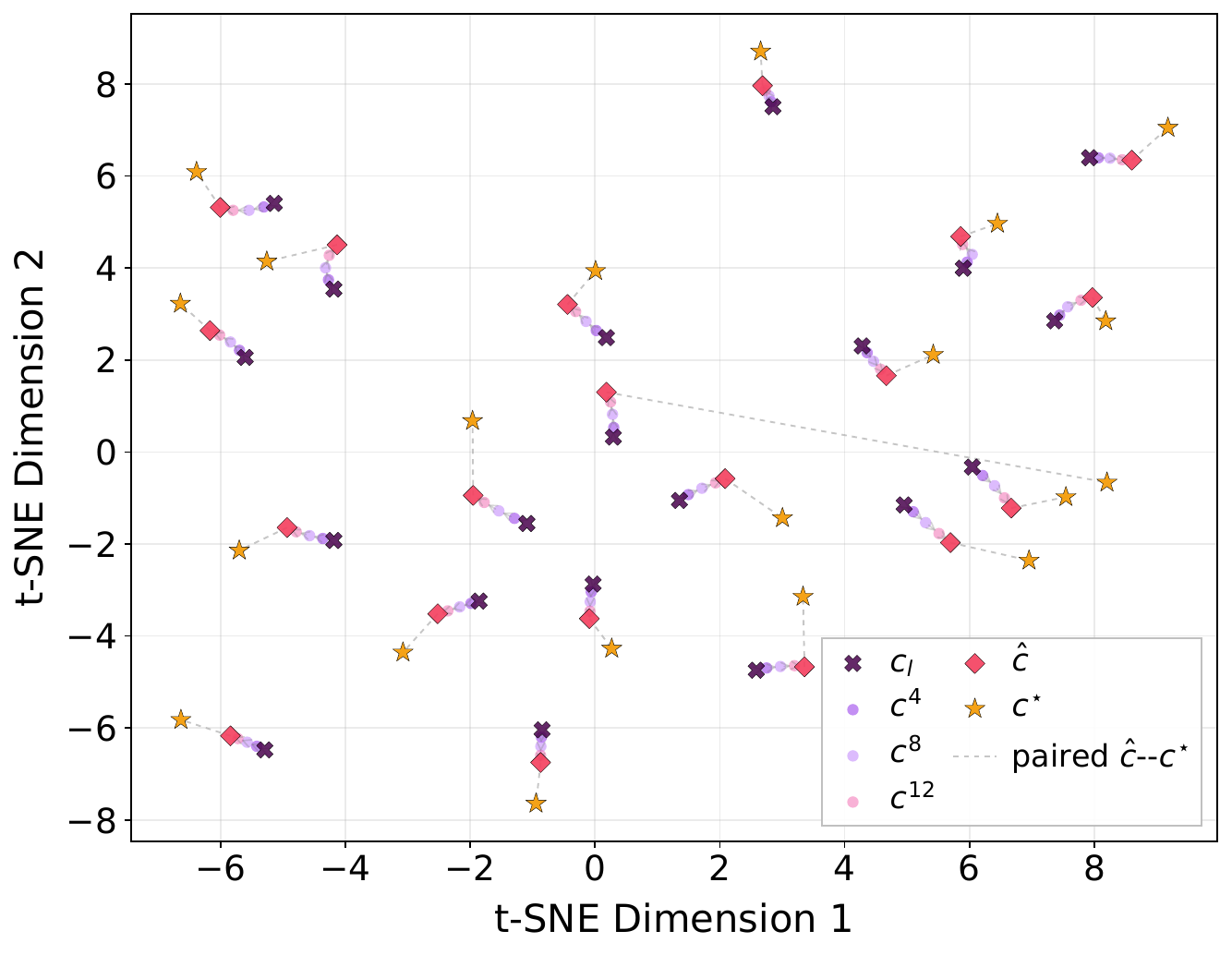}
\vspace{-6mm}
\caption{FMPR prior recovery trajectory.}
\label{fig:vis_fm}
\vspace{-8mm}
\end{wrapfigure}
A straightforward alternative is to directly regress $c^{\star}$ from $c_l$. However, under severe degradation, the mapping from the observed prior to the privileged prior can be highly ambiguous. 
Motivated by flow-matching generative modeling~\citep{lipman2023flow, liu2022flow}, we formulate prior recovery as a flow-matching transport problem.

Specifically, we learn a velocity field $\mathcal{V}_{\mathrm{FM}}$ over the conditional embedding space and integrate it from the observed prior. For each paired sample $(c_l,c^{\star})$, we define the straight interpolation path:
\begin{equation}
    c(t)
    =
    (1-t)c_l+t c^{\star},
    \quad
    \mathcal{V}_{\mathrm{FM}}(c(t),t)
    =
    \frac{d c(t)}{d t}
    = 
    c^{\star} - c_l.
    \qquad
    \label{eq:lbm_compact}
\end{equation}
Because the latent space is highly compact, we integrate Eq.~\eqref{eq:lbm_compact} using $K$ Euler steps for both training and inference. Specifically, we apply:
\begin{equation}
    c^{k+1}=c^k+\frac{1}{K}\mathcal{V}_{\mathrm{FM}}\!\left(c^k,\frac{k}{K}\right),
    \label{eq:K_step}
\end{equation}
starting from $c^0=c_l$ and obtaining the recovered prior $\hat{c}=c^K$, as visualized for 20 representative samples in Fig.~\ref{fig:vis_fm}, where \(c_l\), \(\hat{c}\), \(c^\star\) and intermediate states \(\{c^4,c^8,c^{12}\}\) are projected into a 2D t-SNE space. Then,  \(\hat{c}\) is used as the text-aware condition for restoration: $\hat{z}^{\mathrm{r}}_h=\mathcal{U}_{\theta_{\mathrm{r}}}(z_l,\hat{c}), \hat{x}^{\mathrm{r}}=\mathcal{D}_{\mathrm{vae}}(\hat{z}^{\mathrm{r}}_h),$ where $\mathcal{U}_{\theta_{\mathrm{r}}}$ is initialized from the privileged-prior backbone $\mathcal{U}_{\theta_{\mathrm{p}}}$ and further adapted with the recovered prior. The objective combines image-level restoration supervision and latent prior matching:
\begin{equation}
    \mathcal{L}_{\mathrm{stage1}}
    =
    \underbrace{
    \|\hat{x}^{\mathrm{r}}-x_h\|_1
    +
    \lambda_{\mathrm{lpips}}
    \mathcal{L}_{\mathrm{LPIPS}}(\hat{x}^{\mathrm{r}},x_h)
    }_{\mathcal{L}_{\mathrm{img}}}
    +
    \lambda_{\mathrm{fm}}
    \underbrace{
    \|\hat{c}-c^{\star}\|_1
    }_{\mathcal{L}_{\mathrm{fm}}}.
    \label{eq:stage1_loss_compact}
\end{equation}
This stage stabilizes the high-level text-aware condition under severe degradation, guiding the model toward plausible character identities and coarse structures. However, the recovered prior is still an embedding-space condition, which does not explicitly determine where uncertain local stroke boundaries should be placed in the image. This motivates the next stage, which performs explicit structure refinement under the recovered prior.

\vspace{-2mm}
\subsection{SURE: Structure-guided Uncertainty-aware Residual Encoder}
\vspace{-2mm}
\label{sec:structure_control}
FMPR learned in Sec.~\ref{sec:recoverable_prior} stabilizes global text identity, but local stroke boundaries can still be ambiguous. To address this, after the recoverable prior is learned, we freeze the recovered-prior pathway and the backbone, and train a structure-guided uncertainty-aware residual encoder (SURE). Specifically, SURE consists of two cascading modules: an uncertainty-aware spatial cue extractor $\mathcal{F}_{\eta}$ and a structural residual encoder $\mathcal{C}_{\eta}$. SURE
focuses exclusively on local structural correction.

\vspace{-2mm}
\paragraph{Uncertainty-Aware Spatial Cue Extraction.}
The degraded input contains partial but unevenly reliable structural evidence. Since LQ-derived edges may be incomplete or misleading, treating them as deterministic constraints can amplify degradation artifacts or hallucinate incorrect boundaries. We therefore model the spatial cue in an uncertainty-aware manner, following the general practice of uncertainty-aware prediction for ambiguous visual evidence~\citep{kendall2017uncertainties, NEURIPS2021_88a19961, fang2023UFPNet}.

A spatial cue extractor $\mathcal{F}_{\eta}$ first produces a feature map $f=\mathcal{F}_{\eta}(x_l)$. From $f$, two lightweight heads predict the mean and log-variance of a latent structural cue distribution, denoted as $\mu=h_{\mu}(f)$ and $\log\sigma^2=h_{\sigma}(f)$. We then sample a stochastic structural cue via reparameterization:
\vspace{-1mm}
\begin{equation}
    z_s
    =
    \mu+\sigma\odot\epsilon,
    \qquad
    \epsilon\sim\mathcal{N}(0,I),
    \qquad
    \sigma=\exp\left(\frac{1}{2}\log\sigma^2\right).
    \label{eq:structure_reparam_compact}
\end{equation}
Compared with a deterministic cue, this formulation allows ambiguous regions to be represented with higher uncertainty instead of forcing all local evidence into a single confident estimate. The sampled cue $z_s$ is projected into the structure control space as $p_s=\Pi(z_s)$, and simultaneously decoded by an edge head into an auxiliary boundary map $\hat{m}=h_m(z_s)$ for loss regulation. 

\vspace{-1.5mm}
\paragraph{Structure Control Branch.}
Let $\hat{c}$ denote the recovered prior in Sec.~\ref{sec:recoverable_prior}. 
Given the degraded latent $z_l$, recovered prior $\hat{c}$, and projected structural cue $p_s$, the structure control branch predicts residual signals $\mathcal{R}$ that are then injected into the skip-connection features of the frozen UNet $\mathcal{U}_{\bar{\theta}}$:

\begin{equation}
    \mathcal{R}
    =
    \{r_i\}_{i=1}^{M}
    =
    \mathcal{C}_{\eta}
    \left(
        z_l,\hat{c},p_s
    \right),
    \qquad
    \hat{z}^{\mathrm{s}}_h
    =
    \mathcal{U}_{\bar{\theta}}
    \left(
        z_l,\hat{c};\mathcal{R}
    \right),
    \qquad
    \hat{x}^{\mathrm{s}}
    =
    \mathcal{D}_{\mathrm{vae}}(\hat{z}^{\mathrm{s}}_h),
    \label{eq:structure_control_compact}
\end{equation}
where $\mathcal{C}_{\eta}$ is the structural residual encoder and is encouraged to improve restoration through spatial refinement rather than by re-estimating the text-aware condition.

In practice, we implement $\mathcal{C}_{\eta}$ by initializing its architecture and weights from the diffusion backbone's encoder for simplicity. This allows image-space structural cues to be injected into multiple layers of the frozen backbone while preserving the prior-guided capability learned in the previous stage.

\paragraph{Training objective.}
\begin{wrapfigure}[12]{r}{0.5\columnwidth}
    \vspace{-5mm}%
    \centering
    \setlength{\tabcolsep}{0pt}
    \renewcommand{\arraystretch}{0}

    \begin{adjustbox}{max width=\linewidth}
    \begin{tabular}{@{}r@{\hspace{1mm}}c@{}}

    \raisebox{-.5\height}{\makebox[0.3\linewidth][r]{\large \shortstack[r]{LQ}}} &
    \raisebox{-.5\height}{\includegraphics[height=0.22\linewidth,keepaspectratio]{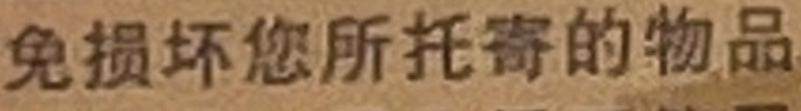}}
    \\[2pt]

    \raisebox{-.5\height}{\makebox[0.3\linewidth][r]{\large \shortstack[r]{LQ-derived\\Boundary Map}}} &
    \raisebox{-.5\height}{\includegraphics[height=0.22\linewidth,keepaspectratio]{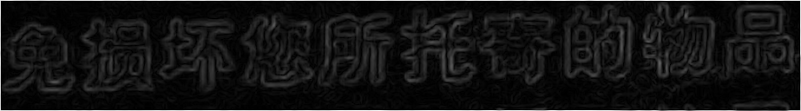}}
    \\[2pt]

    \raisebox{-.5\height}{\makebox[0.3\linewidth][r]{\large \shortstack[r]{Uncertainty\\Map $\sigma$}}} &
    \raisebox{-.5\height}{\includegraphics[height=0.22\linewidth,keepaspectratio]{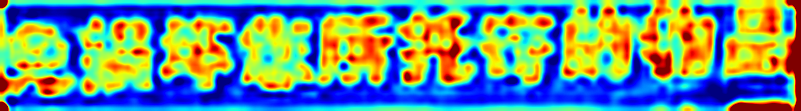}}
    \\[2pt]

    \raisebox{-.5\height}{\makebox[0.3\linewidth][r]{\large \shortstack[r]{Boundary\\Map $\hat{m}$}}} &
    \raisebox{-.5\height}{\includegraphics[height=0.22\linewidth,keepaspectratio]{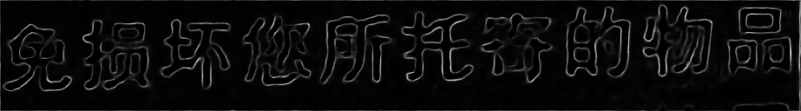}}
    \\[2pt]

    \raisebox{-.5\height}{\makebox[0.3\linewidth][r]{\large \shortstack[r]{Boundary\\Target $m_h$}}} &
    \raisebox{-.5\height}{\includegraphics[height=0.22\linewidth,keepaspectratio]{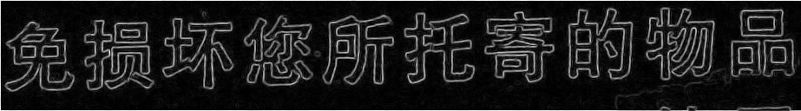}}

    \end{tabular}
    \end{adjustbox}

    \vspace{-1mm}%
    \caption{SURE structural cue visualization.}
    \label{fig:vis_sure}
    \vspace{-10pt}
\end{wrapfigure}
To ensure that the structure branch learns meaningful stroke-level refinement rather than arbitrary feature perturbation, we impose explicit structure-aware supervision. We use the Sobel operator to extract a boundary target $m_h=\mathcal{S}(x_h)$ from the clean image. We further impose a KL penalty between the predicted latent distribution and a standard Gaussian prior. This prevents the variance from collapsing to zero or becoming arbitrarily unstable, thereby preserving the uncertainty-aware nature of the structural cue. The full objective for structure control is:
\begin{equation}
    \mathcal{L}_{\mathrm{stage2}}
    =
    \underbrace{
    \|\hat{x}^{\mathrm{s}}-x_h\|_1
    +
    \lambda_{\mathrm{lpips}}
    \mathcal{L}_{\mathrm{LPIPS}}(\hat{x}^{\mathrm{s}},x_h)
    }_{\mathcal{L}_{\mathrm{img}}}
    + \lambda_{\mathrm{str}}
    \underbrace{\|\hat{m}-m_h\|_1}_{\mathcal{L}_{\mathrm{str}}}
    +
    \lambda_{\mathrm{kl}}
    \underbrace{
    D_{\mathrm{KL}}
    \left(
    \mathcal{N}(\mu,\sigma^2)
    \,\|\, 
    \mathcal{N}(0,I)
    \right)
    }_{\mathcal{L}_{\mathrm{kl}}}.
\end{equation}
As visualized in Fig.~\ref{fig:vis_sure} (LQ, LQ-derived boundary map, uncertainty map $\sigma$, $\hat{m}$, and $m_h$), the model generates distinctly clearer structures in $\hat{m}$ where it exhibits high confidence (i.e., low uncertainty, indicated by the red areas in the uncertainty map), whereas regions with high uncertainty appear correspondingly blurry in $\hat{m}$. By feeding these uncertainty-aware regularized features into $\mathcal{C}_{\eta}$, the model can more effectively focus on local stroke topology, boundary closure, and spatial alignment.

\vspace{-2mm}
\section{Experiments}
\label{sec:experiments}
\vspace{-1mm}
\subsection{Experimental Setup}

\vspace{-2mm}
\paragraph{Datasets.}
We focus on Chinese-English text-line SR. Existing Text-SR datasets differ in language coverage, image quality, scale, and task scope, making it difficult to form a consistent training corpus for this task. TextZoom~\citep{wang2020tsrn} provides real-world English pairs but lacks broader bilingual coverage. Real-CE~\citep{ma2023realce} contains Chinese-English real text pairs, but is relatively limited in scale. SA-Text~\citep{min2026terediff} provides high-quality scene images with dense text annotations, but our corpus analysis shows limited usable Chinese text crops, as detailed in the appendix. We therefore construct BTL by combining filtered real text crops from existing annotated sources with synthetic text-line rendering.

Specifically, we collect Chinese text crops with annotations from the CTR benchmark~\citep{yu2021ctr}, extract English text crops from SA-Text annotations~\citep{min2026terediff}, and include digit samples from both sources. All candidates are filtered by unified criteria: (i) valid annotations; (ii) resized height of 128 pixels; (iii) aspect ratios between 2 and 8; (iv) transcripts no longer than 24 characters; and (v) no-reference IQA-based quality ranking. For quality ranking, we use a weighted score based on MUSIQ~\citep{ke2021musiq}, MANIQA~\citep{yang2022maniqa}, and CLIP-IQA~\citep{wang2023clipiqa}. This process yields 50K quality-controlled real text-line images.

To improve text appearance and layout diversity, we further synthesize 50K high-quality text-line images following the synthetic text rendering strategy of MARCONet~\citep{li2023marconet}. Together with the curated real crops, this forms BTL, a 100K HQ bilingual text-line corpus. For each HQ image, we generate an LQ counterpart using degradation pipelines based on BSRGAN~\citep{zhang2021bsrgan} and Real-ESRGAN~\citep{wang2021realesrgan}. We use 80K image pairs for training and reserve 20K pairs for synthetic evaluation, denoted as BTL-train and BTL-test, respectively.

We further evaluate real-world generalization on RealCE-val. Since some LQ-HQ pairs exhibit noticeable misalignment, color mismatch, or annotation errors, we filter invalid pairs and manually correct erroneous annotations, resulting in 1,037 valid testing pairs. Detailed construction rules, source statistics, and final distributions of BTL are provided in the appendix.

\begin{table}[!t]
\centering
\small
\setlength{\tabcolsep}{3.2pt}
\renewcommand{\arraystretch}{0.95}

\textbf{(a) Synthetic dataset BTL-test.}
\vspace{1mm}

\begin{adjustbox}{width=\linewidth}
\begin{tabular}{l|ccccc|ccccc}
\toprule
\multirow{2}{*}{Methods} & \multicolumn{5}{c|}{$\times 2$} & \multicolumn{5}{c}{$\times 4$} \\
 & PSNR $\uparrow$ & LPIPS $\downarrow$ & FID $\downarrow$ & ACC $\uparrow$ & NED $\uparrow$
 & PSNR $\uparrow$ & LPIPS $\downarrow$ & FID $\downarrow$ & ACC $\uparrow$ & NED $\uparrow$ \\
\midrule
TSRN~\citep{wang2020tsrn}    & 22.28 & 0.3581 & 48.85 & 52.63\% & 0.7363
        & 21.53 & 0.5039 & 98.65 & 31.37\% & 0.5046 \\
TBSRN~\citep{chen2021tbsrn}   & 24.47 & 0.3619 & 65.74 & 51.69\% & 0.7243
        & \underline{22.25} & 0.4865 & 95.46 & 33.41\% & 0.5241 \\
TATT~\citep{ma2022tatt}    & 24.51 & 0.3584 & 57.40 & 50.07\% & 0.7154
        & 20.20 & 0.4978 & 83.73 & 29.80\% & 0.4826 \\
MARCONet~\citep{li2023marconet}    & \textbf{25.03} & \underline{0.1974} & 17.88 & 59.08\% & 0.8075
            & \textbf{22.35} & \underline{0.2956} & 33.48 & \underline{37.96\%} & \underline{0.6049} \\
DiffTSR~\citep{zhang2024difftsr} & 24.21 & 0.2327 & 15.96 & \textbf{59.85\%} & \underline{0.8117}
        & 21.00 & 0.3352 & \underline{25.55} & 37.85\% & 0.6007 \\
StyleSRN~\citep{yuan2025stylesrn}    & 12.44 & 0.7272 & 119.89 & 26.55\% & 0.4242
            & 12.30 & 0.7723 & 142.33 & 19.05\% & 0.3248 \\
TeReDiff~\citep{min2026terediff}    & 22.38 & 0.2692 & \underline{13.22} & 49.70\% & 0.7272
        & 20.16 & 0.3832 & 27.14 & 29.14\% & 0.4910 \\
\midrule
PRISM & \underline{24.53} & \textbf{0.1514} & \textbf{6.14} & \underline{59.78\%} & \textbf{0.8220}
     & 22.08 & \textbf{0.2314} & \textbf{12.57} & \textbf{42.12\%} & \textbf{0.6644} \\
\bottomrule
\end{tabular}
\end{adjustbox}

\vspace{2mm}

\textbf{(b) Real-world dataset RealCE-val.}
\vspace{1mm}

\begin{adjustbox}{width=\linewidth}
\begin{tabular}{l|ccccc|ccccc}
\toprule
\multirow{2}{*}{Methods} & \multicolumn{5}{c|}{$\times 2$} & \multicolumn{5}{c}{$\times 4$} \\
 & PSNR $\uparrow$ & LPIPS $\downarrow$ & FID $\downarrow$ & ACC $\uparrow$ & NED $\uparrow$
 & PSNR $\uparrow$ & LPIPS $\downarrow$ & FID $\downarrow$ & ACC $\uparrow$ & NED $\uparrow$ \\
\midrule
TSRN~\citep{wang2020tsrn}    & 19.51 & 0.1420 & 58.88 & 82.64\% & 0.9392
        & \underline{19.32} & 0.2988 & 141.70 & 52.17\% & 0.7860 \\
TBSRN~\citep{chen2021tbsrn}   & 20.32 & \textbf{0.1343} & 51.66 & 83.99\% & \textbf{0.9480}
        & 19.04 & 0.2767 & 111.07 & 60.17\% & 0.8324 \\
TATT~\citep{ma2022tatt}    & \underline{20.54} & 0.2118 & 80.67 & 80.71\% & 0.9385
        & 18.16 & 0.3048 & 122.78 & 46.67\% & 0.7538 \\
MARCONet~\citep{li2023marconet}    & 19.38 & 0.1704 & 54.03 & \textbf{84.36\%} & 0.9377
            & 18.91 & \underline{0.2326} & \underline{74.52} & \underline{60.62\%} & \underline{0.8381} \\
DiffTSR~\citep{zhang2024difftsr} & 20.26 & 0.1964 & \underline{35.59} & 81.97\% & 0.9424
        & 18.60 & 0.3052 & 74.94 & 54.00\% & 0.8077 \\
StyleSRN~\citep{yuan2025stylesrn}    & 14.12 & 0.5009 & 90.51 & 51.30\% & 0.7043
            & 13.97 & 0.5403 & 101.39 & 38.67\% & 0.6083 \\
TeReDiff~\citep{min2026terediff}    & 17.81 & 0.2913 & 88.15 & 52.27\% & 0.6969
        & 17.03 & 0.3632 & 106.13 & 34.23\% & 0.5675 \\
\midrule
PRISM & \textbf{21.00} & \underline{0.1372} & \textbf{33.71} & \underline{84.28\%} & \underline{0.9442}
     & \textbf{19.89} & \textbf{0.2043} & \textbf{47.83} & \textbf{65.19\%} & \textbf{0.8521} \\
\bottomrule
\end{tabular}
\end{adjustbox}

\vspace{0.5mm}
\caption{Quantitative comparison on BTL-test and RealCE-val under $\times2$ and $\times4$ text image super-resolution. Best and second-best results are shown in bold and underlined, respectively.}
\label{tab:main_results}
\vspace{-9mm}
\end{table}

\vspace{-3mm}
\paragraph{Implementation Details.}
\label{sec:implementation_details}
We build our model on the pretrained Stable Diffusion 2.1-base model and train the UNet with LoRA\citep{hu2022lora} of rank 16. FMPR contains two training stages, privileged-prior construction and LQ-only prior recovery, each trained for 100K iterations. SURE is then trained for 50K iterations with the FMPR pathway and restoration backbone frozen. All stages use AdamW with a learning rate of \(5\times10^{-5}\) and a total batch size of 8 on two NVIDIA RTX A6000 GPUs. FMPR uses 16-step Euler discretization in Eq.~\eqref{eq:K_step} to recover the text prior, and final restoration is performed in one step at a fixed timestep of 399~\citep{li2024distillation,wang2025osdface}. We set \(\lambda_{\mathrm{lpips}}=1\) and \(\lambda_{\mathrm{fm}}=1\) for FMPR, and \(\lambda_{\mathrm{lpips}}=1\), \(\lambda_{\mathrm{str}}=1\), and \(\lambda_{\mathrm{kl}}=0.1\) for SURE.

\vspace{-3mm}
\paragraph{Compared Methods and Evaluation Metrics.}
We compare our method with representative Text-SR methods, including TSRN~\citep{wang2020tsrn}, TBSRN~\citep{chen2021tbsrn}, TATT~\citep{ma2022tatt}, MARCONet~\citep{li2023marconet}, DiffTSR~\citep{zhang2024difftsr}, and StyleSRN~\citep{yuan2025stylesrn}. We also include TeReDiff~\citep{min2026terediff}, a recent text-aware image restoration method. For fair comparison, all trainable baselines are retrained or fine-tuned on BTL-train following their official settings. We evaluate reconstruction fidelity with peak signal-to-noise ratio (PSNR) and learned perceptual image patch similarity (LPIPS)~\citep{zhang2018lpips}, which measure image-space and feature-space differences from the reference image, respectively. We use Fr\'echet inception distance (FID)~\citep{heusel2017fid} to assess distributional realism. For text fidelity, we report OCR accuracy (ACC) and normalized edit distance (NED)~\citep{ma2023realce}, both computed using PP-OCRv5~\citep{cui2025paddleocr} as the recognition model.

\vspace{-3mm}
\subsection{Main Results}
\vspace{-2mm}
\paragraph{Quantitative Comparisons.}
Table~\ref{tab:main_results} reports quantitative comparisons on BTL-test and RealCE-val under $\times2$ and $\times4$ settings. On BTL-test, our method obtains the best LPIPS, FID, and NED at both scales, and the highest ACC under $\times4$. These results show clear advantages in perceptual quality and text fidelity. Although there is a gap in PSNR, it aligns with the well-known perception-distortion tradeoff: unlike PSNR-oriented methods that tend to produce overly smoothed outputs, our diffusion-based approach recovers sharp, high-frequency stroke details that significantly benefit character readability. While our model is trained on BTL-train, it also generalizes well to real-world degraded text images. On RealCE-val, our method achieves the best PSNR and FID and ranks second in the remaining metrics under $\times2$, and ranks first across all metrics under the more challenging $\times4$ setting. Notably, under $\times4$, it improves ACC from 60.62\% to 65.19\% and reduces FID from 74.52 to 47.83 compared with the second-best results. Overall, the results show that the proposed method improves perceptual realism and character-level readability, especially under severe real-world degradation.

\begin{figure*}[t]
\centering

\setlength{\tabcolsep}{0pt}
\renewcommand{\arraystretch}{0}
\hspace{-15mm}
\begin{adjustbox}{max width=1.1\textwidth}
\begin{tabular}{@{}r@{\hspace{4mm}}c@{\hspace{1mm}}c@{\hspace{1mm}}c@{\hspace{1mm}}c@{}}

\raisebox{-.5\height}{\makebox[0.2\textwidth][r]{\large GT}} &
\raisebox{-.5\height}{\includegraphics[height=0.1\textwidth,keepaspectratio]{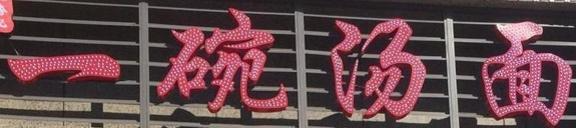}} &
\raisebox{-.5\height}{\includegraphics[height=0.1\textwidth,keepaspectratio]{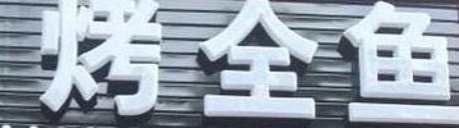}} &
\raisebox{-.5\height}{\includegraphics[height=0.1\textwidth,keepaspectratio]{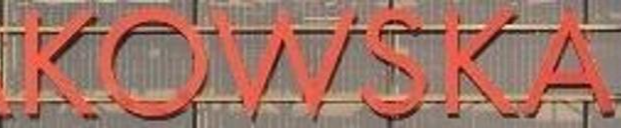}} &
\raisebox{-.5\height}{\includegraphics[height=0.1\textwidth,keepaspectratio]{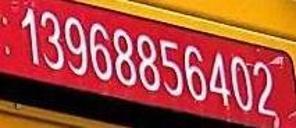}}
\vspace{1mm}
\\[2pt]

\raisebox{-.5\height}{\makebox[0.2\textwidth][r]{\large LR}} &
\raisebox{-.5\height}{\includegraphics[height=0.1\textwidth,keepaspectratio]{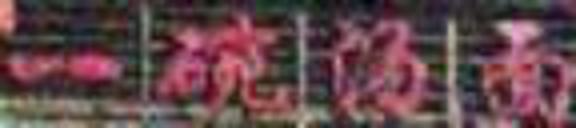}} &
\raisebox{-.5\height}{\includegraphics[height=0.1\textwidth,keepaspectratio]{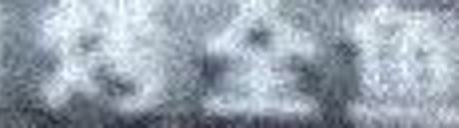}} &
\raisebox{-.5\height}{\includegraphics[height=0.1\textwidth,keepaspectratio]{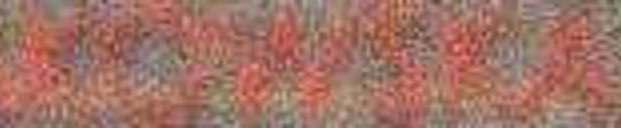}} &
\raisebox{-.5\height}{\includegraphics[height=0.1\textwidth,keepaspectratio]{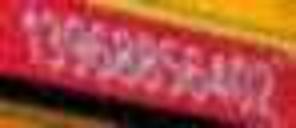}}
\vspace{1mm}
\\[2pt]

\raisebox{-.5\height}{\makebox[0.2\textwidth][r]{\large TATT}} &
\raisebox{-.5\height}{\includegraphics[height=0.1\textwidth,keepaspectratio]{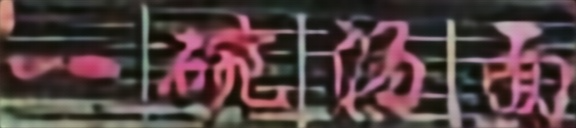}} &
\raisebox{-.5\height}{\includegraphics[height=0.1\textwidth,keepaspectratio]{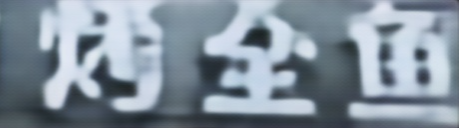}} &
\raisebox{-.5\height}{\includegraphics[height=0.1\textwidth,keepaspectratio]{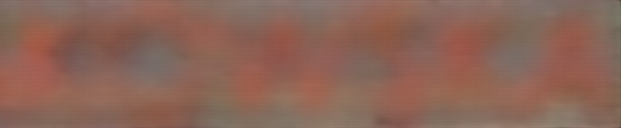}} &
\raisebox{-.5\height}{\includegraphics[height=0.1\textwidth,keepaspectratio]{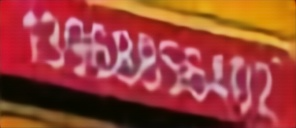}}
\vspace{1mm}
\\[2pt]

\raisebox{-.5\height}{\makebox[0.2\textwidth][r]{\large StyleSRN}} &
\raisebox{-.5\height}{\includegraphics[height=0.1\textwidth,keepaspectratio]{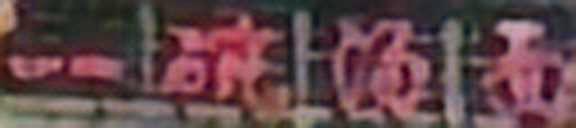}} &
\raisebox{-.5\height}{\includegraphics[height=0.1\textwidth,keepaspectratio]{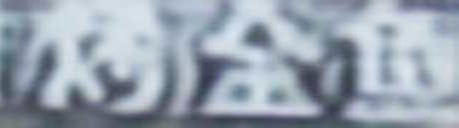}} &
\raisebox{-.5\height}{\includegraphics[height=0.1\textwidth,keepaspectratio]{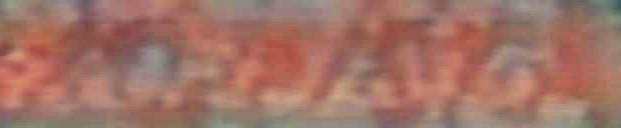}} &
\raisebox{-.5\height}{\includegraphics[height=0.1\textwidth,keepaspectratio]{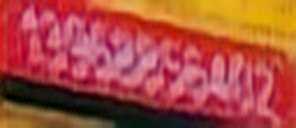}}
\vspace{1mm}
\\[2pt]

\raisebox{-.5\height}{\makebox[0.2\textwidth][r]{\large MARCONet}} &
\raisebox{-.5\height}{\includegraphics[height=0.1\textwidth,keepaspectratio]{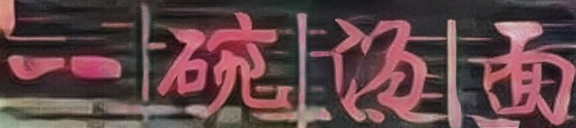}} &
\raisebox{-.5\height}{\includegraphics[height=0.1\textwidth,keepaspectratio]{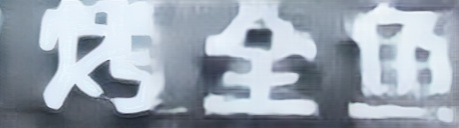}} &
\raisebox{-.5\height}{\includegraphics[height=0.1\textwidth,keepaspectratio]{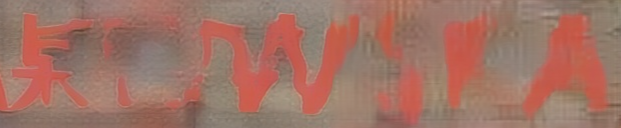}} &
\raisebox{-.5\height}{\includegraphics[height=0.1\textwidth,keepaspectratio]{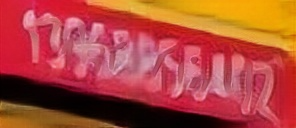}}
\vspace{1mm}
\\[2pt]

\raisebox{-.5\height}{\makebox[0.2\textwidth][r]{\large DiffTSR}} &
\raisebox{-.5\height}{\includegraphics[height=0.1\textwidth,keepaspectratio]{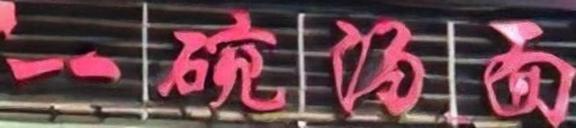}} &
\raisebox{-.5\height}{\includegraphics[height=0.1\textwidth,keepaspectratio]{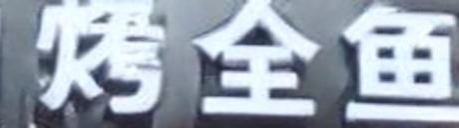}} &
\raisebox{-.5\height}{\includegraphics[height=0.1\textwidth,keepaspectratio]{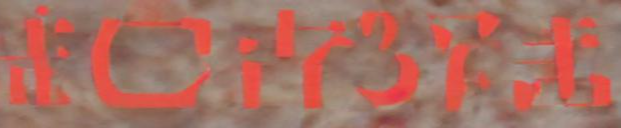}} &
\raisebox{-.5\height}{\includegraphics[height=0.1\textwidth,keepaspectratio]{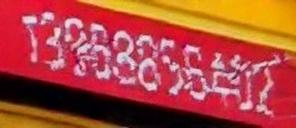}}
\vspace{1mm}
\\[2pt]

\raisebox{-.5\height}{\makebox[0.2\textwidth][r]{\large TeReDiff}} &
\raisebox{-.5\height}{\includegraphics[height=0.1\textwidth,keepaspectratio]{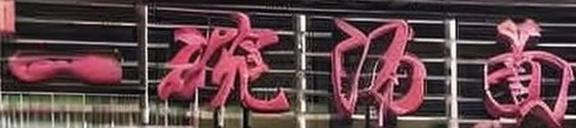}} &
\raisebox{-.5\height}{\includegraphics[height=0.1\textwidth,keepaspectratio]{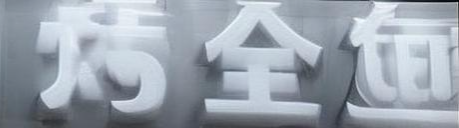}} &
\raisebox{-.5\height}{\includegraphics[height=0.1\textwidth,keepaspectratio]{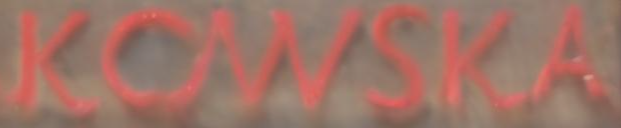}} &
\raisebox{-.5\height}{\includegraphics[height=0.1\textwidth,keepaspectratio]{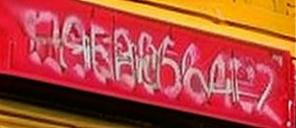}}
\vspace{1mm}
\\[2pt]

\raisebox{-.5\height}{\makebox[0.2\textwidth][r]{\large PRISM}} &
\raisebox{-.5\height}{\includegraphics[height=0.1\textwidth,keepaspectratio]{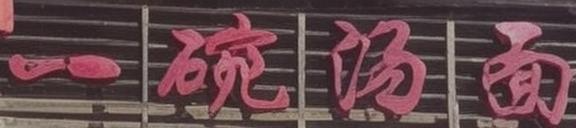}} &
\raisebox{-.5\height}{\includegraphics[height=0.1\textwidth,keepaspectratio]{main/btl/sample2/9_ours.png}} &
\raisebox{-.5\height}{\includegraphics[height=0.1\textwidth,keepaspectratio]{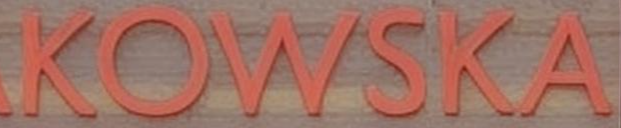}} &
\raisebox{-.5\height}{\includegraphics[height=0.1\textwidth,keepaspectratio]{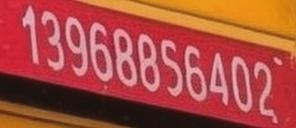}}

\end{tabular}
\end{adjustbox}
\vspace{-2mm}
\caption{Qualitative comparison on the synthetic BTL-test dataset for $\times4$ super-resolution. We compare our method with TATT~\citep{ma2022tatt}, StyleSRN~\citep{yuan2025stylesrn}, MARCONet~\citep{li2023marconet}, DiffTSR~\citep{zhang2024difftsr}, and TeReDiff~\citep{min2026terediff}.}
\label{fig:qual_btl}
\vspace{-3mm}
\end{figure*}

\begin{figure*}[t]
\centering

\setlength{\tabcolsep}{0pt}
\renewcommand{\arraystretch}{0}
\hspace{-15mm}
\begin{adjustbox}{max width=1.1\textwidth}
\begin{tabular}{@{}r@{\hspace{4mm}}c@{\hspace{1mm}}c@{\hspace{1mm}}c@{\hspace{1mm}}c@{}}

\raisebox{-.5\height}{\makebox[0.2\textwidth][r]{\large GT}} &
\raisebox{-.5\height}{\includegraphics[height=0.1\textwidth,keepaspectratio]{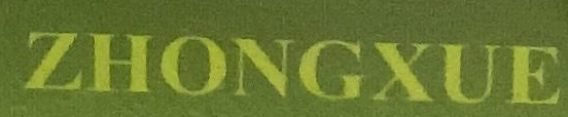}} &
\raisebox{-.5\height}{\includegraphics[height=0.1\textwidth,keepaspectratio]{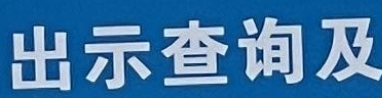}} &
\raisebox{-.5\height}{\includegraphics[height=0.1\textwidth,keepaspectratio]{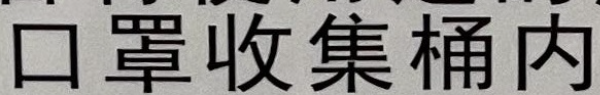}} &
\raisebox{-.5\height}{\includegraphics[height=0.1\textwidth,keepaspectratio]{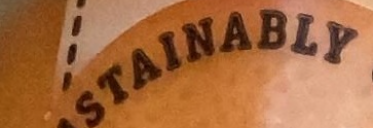}}
\vspace{1mm}
\\[2pt]

\raisebox{-.5\height}{\makebox[0.2\textwidth][r]{\large LR}} &
\raisebox{-.5\height}{\includegraphics[height=0.1\textwidth,keepaspectratio]{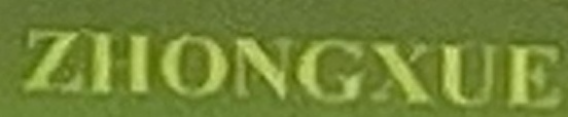}} &
\raisebox{-.5\height}{\includegraphics[height=0.1\textwidth,keepaspectratio]{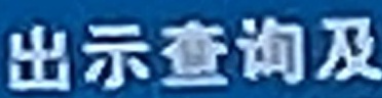}} &
\raisebox{-.5\height}{\includegraphics[height=0.1\textwidth,keepaspectratio]{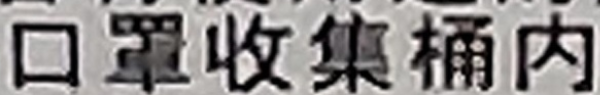}} &
\raisebox{-.5\height}{\includegraphics[height=0.1\textwidth,keepaspectratio]{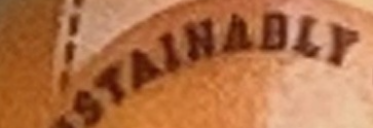}}
\vspace{1mm}
\\[2pt]

\raisebox{-.5\height}{\makebox[0.2\textwidth][r]{\large TATT}} &
\raisebox{-.5\height}{\includegraphics[height=0.1\textwidth,keepaspectratio]{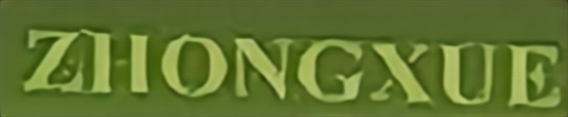}} &
\raisebox{-.5\height}{\includegraphics[height=0.1\textwidth,keepaspectratio]{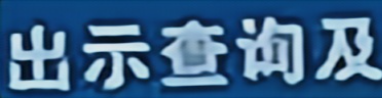}} &
\raisebox{-.5\height}{\includegraphics[height=0.1\textwidth,keepaspectratio]{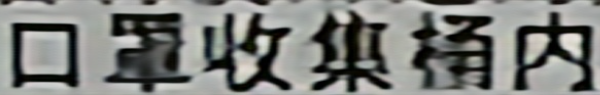}} &
\raisebox{-.5\height}{\includegraphics[height=0.1\textwidth,keepaspectratio]{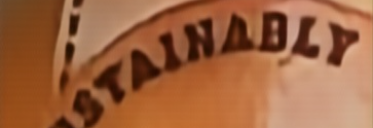}}
\vspace{1mm}
\\[2pt]

\raisebox{-.5\height}{\makebox[0.2\textwidth][r]{\large StyleSRN}} &
\raisebox{-.5\height}{\includegraphics[height=0.1\textwidth,keepaspectratio]{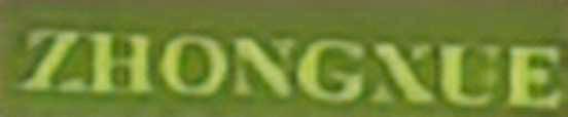}} &
\raisebox{-.5\height}{\includegraphics[height=0.1\textwidth,keepaspectratio]{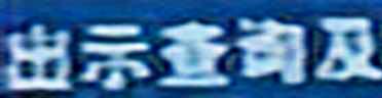}} &
\raisebox{-.5\height}{\includegraphics[height=0.1\textwidth,keepaspectratio]{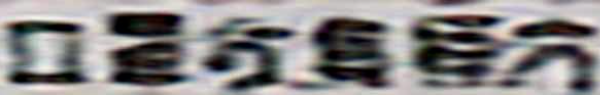}} &
\raisebox{-.5\height}{\includegraphics[height=0.1\textwidth,keepaspectratio]{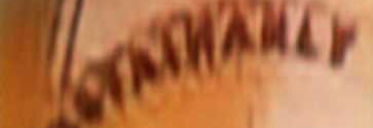}}
\vspace{1mm}
\\[2pt]

\raisebox{-.5\height}{\makebox[0.2\textwidth][r]{\large MARCONet}} &
\raisebox{-.5\height}{\includegraphics[height=0.1\textwidth,keepaspectratio]{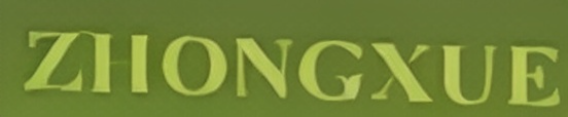}} &
\raisebox{-.5\height}{\includegraphics[height=0.1\textwidth,keepaspectratio]{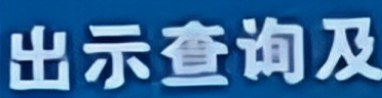}} &
\raisebox{-.5\height}{\includegraphics[height=0.1\textwidth,keepaspectratio]{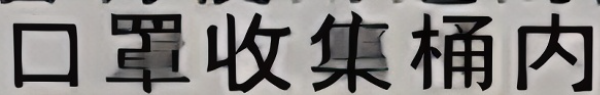}} &
\raisebox{-.5\height}{\includegraphics[height=0.1\textwidth,keepaspectratio]{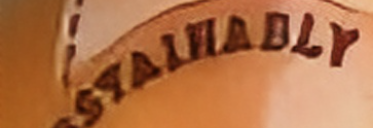}}
\vspace{1mm}
\\[2pt]

\raisebox{-.5\height}{\makebox[0.2\textwidth][r]{\large DiffTSR}} &
\raisebox{-.5\height}{\includegraphics[height=0.1\textwidth,keepaspectratio]{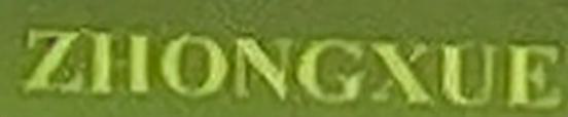}} &
\raisebox{-.5\height}{\includegraphics[height=0.1\textwidth,keepaspectratio]{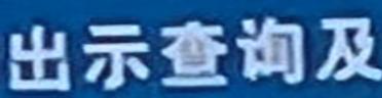}} &
\raisebox{-.5\height}{\includegraphics[height=0.1\textwidth,keepaspectratio]{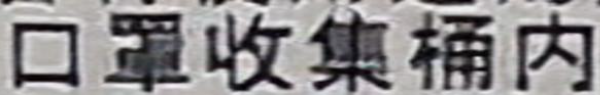}} &
\raisebox{-.5\height}{\includegraphics[height=0.1\textwidth,keepaspectratio]{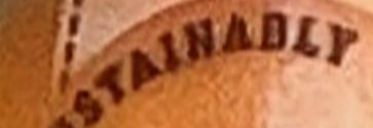}}
\vspace{1mm}
\\[2pt]

\raisebox{-.5\height}{\makebox[0.2\textwidth][r]{\large TeReDiff}} &
\raisebox{-.5\height}{\includegraphics[height=0.1\textwidth,keepaspectratio]{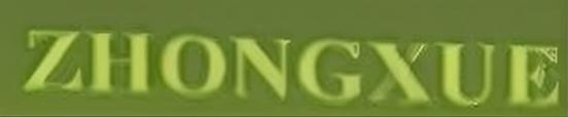}} &
\raisebox{-.5\height}{\includegraphics[height=0.1\textwidth,keepaspectratio]{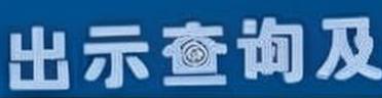}} &
\raisebox{-.5\height}{\includegraphics[height=0.1\textwidth,keepaspectratio]{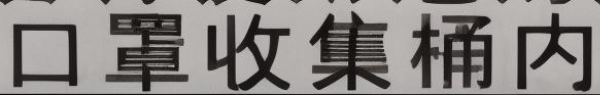}} &
\raisebox{-.5\height}{\includegraphics[height=0.1\textwidth,keepaspectratio]{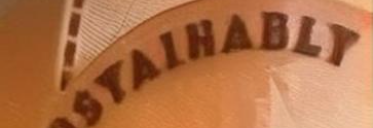}}
\vspace{1mm}
\\[2pt]

\raisebox{-.5\height}{\makebox[0.2\textwidth][r]{\large PRISM}} &
\raisebox{-.5\height}{\includegraphics[height=0.1\textwidth,keepaspectratio]{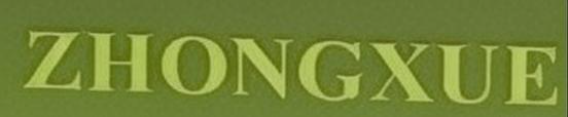}} &
\raisebox{-.5\height}{\includegraphics[height=0.1\textwidth,keepaspectratio]{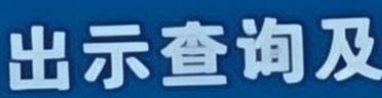}} &
\raisebox{-.5\height}{\includegraphics[height=0.1\textwidth,keepaspectratio]{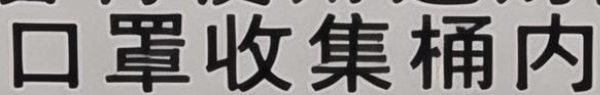}} &
\raisebox{-.5\height}{\includegraphics[height=0.1\textwidth,keepaspectratio]{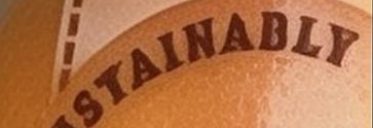}}

\end{tabular}
\end{adjustbox}
\vspace{-2mm}
\caption{Qualitative comparison on the real-world RealCE-val dataset for $\times4$ super-resolution. We compare our method with TATT~\citep{ma2022tatt}, StyleSRN~\citep{yuan2025stylesrn}, MARCONet~\citep{li2023marconet}, DiffTSR~\citep{zhang2024difftsr}, and TeReDiff~\citep{min2026terediff}.}
\label{fig:qual_realce}
\vspace{-4mm}
\end{figure*}

\vspace{-3mm}
\paragraph{Qualitative Comparisons.}
Figures~\ref{fig:qual_btl} and~\ref{fig:qual_realce} compare visual results on BTL-test and RealCE-val. As can be seen, TATT~\citep{ma2022tatt} and StyleSRN~\citep{yuan2025stylesrn} tend to produce over-smoothed text, especially for complex Chinese glyphs under severe blur. MARCONet~\citep{li2023marconet} restores sharper strokes in some cases, but often introduces structural distortion or weak text-background consistency, as shown in the 2nd and 3rd BTL-test examples. Diffusion-based methods improve perceptual sharpness, but still suffer from text-specific artifacts. DiffTSR~\citep{zhang2024difftsr} produces broken or merged strokes under severe degradation, as shown in the 4th BTL-test and 2nd RealCE-val examples. TeReDiff~\citep{min2026terediff} is prone to false-color artifacts on small text images and may hallucinate redundant or incorrect strokes, as shown in the 4th BTL-test and 3rd RealCE-val examples. In contrast, by combining recovered text priors with uncertainty-aware structural cues, our method better preserves character readability and local stroke continuity while maintaining more consistent background appearance.

\vspace{-3mm}
\paragraph{Inference Efficiency.}
Inference efficiency is important for practical Text-SR, especially for diffusion models. Our PRISM uses only one denoising step, compared with 200 steps for DiffTSR~\citep{zhang2024difftsr} and 50 for TeReDiff~\citep{min2026terediff}. We compare speed on test images resized to 128$\times$512. For a single image, our method takes 0.08 s, compared with 10.70 s for DiffTSR and 5.27 s for TeReDiff. Importantly, this single-step design makes our inference speed highly comparable to standard CNN- and Transformer-based methods. Detailed runtime comparisons of all evaluated methods are provided in the appendix.

\begin{figure}[!t]
\centering
    \begin{minipage}{.68\linewidth}
        \centering
        \begin{adjustbox}{width=\linewidth}
        \begin{tabular}{l|cccccc}
        \toprule
        Methods & PSNR $\uparrow$ & LPIPS $\downarrow$ & DISTS $\downarrow$ & FID $\downarrow$ & ACC $\uparrow$ & NED $\uparrow$ \\
        \midrule
        Base model        & 19.8113 & 0.2121 & 0.2143 & 53.42 & 62.39\% & 0.8390 \\
        Privileged reference ($c^{\star}$) & 20.7509 & 0.1211 & 0.1604 & 31.49 & 64.90\% & 0.8588 \\
        \midrule
        Direct regression & 19.8167 & 0.2107 & 0.2123 & 50.67 & 64.03\% & 0.8446 \\
        Diffusion & \textbf{19.9202} & 0.2123 & 0.2109 & 49.40 & 62.49\% & 0.8371 \\
        Flow Matching     & 19.8257 & \textbf{0.2059} & \textbf{0.2097} & \textbf{47.54} & \textbf{64.61\%} & \textbf{0.8461} \\
        \bottomrule
        \end{tabular}
        \end{adjustbox}
        \vspace{-0.5mm}
        \captionof{table}{Ablation study of prior learning paradigms on RealCE-val.}
        \label{tab:ablation_prior_paradigm}
    \end{minipage}
    \hspace{0mm}
    \begin{minipage}{.3\linewidth}
        \centering
        \vspace{-3mm}
        \includegraphics[width=\linewidth, height=0.6\linewidth]{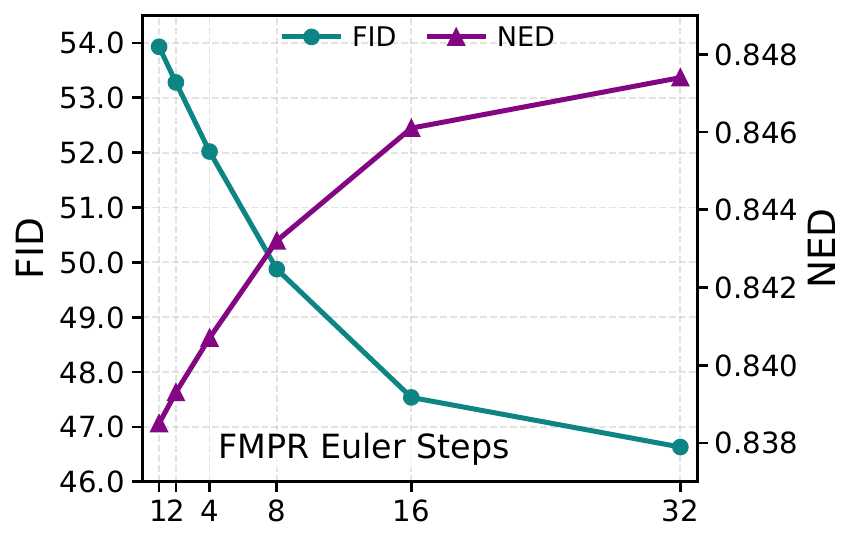}
        \vspace{-6mm}
        \captionof{figure}{FMPR Euler steps.}
        \vspace{-1.5mm}
        \label{fig:ablation_euler_steps}
    \end{minipage}
\vspace{-6mm}
\end{figure}

\vspace{-3mm}
\subsection{Ablation Studies}
\vspace{-2mm}
\noindent \textbf{Analysis of Prior Learning Paradigms.}
We compare different paradigms for recovering the privileged prior from degraded observations on RealCE-val. As shown in Tab.~\ref{tab:ablation_prior_paradigm}, using the privileged condition $c^{\star}$ provides a clear upper bound, confirming that paired LQ/HQ latents define an informative text-aware prior space. The remaining variants examine how such a prior can be approximated from the degraded input alone. Direct regression yields moderate gains but struggles to close the prior gap, as it treats recovery merely as target fitting, which is insufficient to close the gap between degraded and privileged priors. While the diffusion-based variant improves reconstruction, its gains in recognition-oriented metrics remain limited. This suggests that the traditional diffusion approach, constructing the prior from a pure Gaussian distribution under the LQ condition, entails overly complex and redundant generation paths. This unnecessary complexity hinders the effective learning of strict character structures. In contrast, Flow Matching starts from the observed prior and learns a continuous transport field toward the privileged prior space. This effectively rectifies unreliable information while preserving the degraded condition, achieving the strongest overall balance of character-level fidelity and perceptual quality. 

\begin{figure*}[t]
\centering
\hspace{-3.5mm}%
\raisebox{-1.0em}{%
    \begin{minipage}[t]{0.62\textwidth}
        \vspace{0pt}
        \centering
        \resizebox{\linewidth}{!}{
        \begin{tabular}{l|ccccc}
        \toprule
        Methods & PSNR $\uparrow$ & SSIM $\uparrow$ & LPIPS $\downarrow$ & ACC $\uparrow$ & NED $\uparrow$ \\
        \midrule
        FMPR only & 19.8257 & 0.5968 & 0.2059 & 64.61\% & 0.8461 \\
        Residual branch only & 19.8671 & 0.5947 & 0.2099 & 64.80\% & 0.8468 \\
        w/o uncertainty & 19.8768 & 0.5974 & 0.2056 & 64.80\% & 0.8477 \\
        Full model & \textbf{19.8919} & \textbf{0.6012} & \textbf{0.2043} & \textbf{65.19\%} & \textbf{0.8521} \\
        \bottomrule
        \end{tabular}
        }
        
        \captionof{table}{Ablation study of SURE on RealCE-val.}
        \label{tab:ablation_sure}
    \end{minipage}
    }
\hspace{5mm}%
\raisebox{0.8em}{%
    \begin{minipage}[t]{0.33\textwidth}
        \vspace{0pt}
        \centering
        
        \setlength{\tabcolsep}{0pt}
        \renewcommand{\arraystretch}{0}
        
        \newcommand{\labw}{0.005\textwidth}
        \newcommand{\imgh}{0.15\textwidth}
        \newcommand{\cgap}{0.25mm}
        
        \begin{tabular}{@{}r@{\hspace{0.3mm}}c@{\hspace{\cgap}}c@{\hspace{\cgap}}c@{\hspace{\cgap}}c@{}}
        
        \raisebox{-.5\height}{\makebox[\labw][r]{\scriptsize LR}} &
        \raisebox{-.5\height}{\includegraphics[height=\imgh,keepaspectratio]{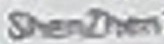}} &
        \raisebox{-.5\height}{\includegraphics[height=\imgh,keepaspectratio]{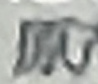}} &
        \raisebox{-.5\height}{\includegraphics[height=\imgh,keepaspectratio]{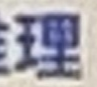}} &
        \raisebox{-.5\height}{\includegraphics[height=\imgh,keepaspectratio]{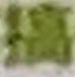}}
        \vspace{\cgap}
        \\
        
        \raisebox{-.5\height}{\makebox[\labw][r]{\scriptsize \shortstack[r]{Base\\Model}}} &
        \raisebox{-.5\height}{\includegraphics[height=\imgh,keepaspectratio]{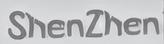}} &
        \raisebox{-.5\height}{\includegraphics[height=\imgh,keepaspectratio]{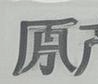}} &
        \raisebox{-.5\height}{\includegraphics[height=\imgh,keepaspectratio]{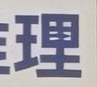}} &
        \raisebox{-.5\height}{\includegraphics[height=\imgh,keepaspectratio]{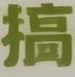}}
        \vspace{\cgap}
        \\
        
        \raisebox{-.5\height}{\makebox[\labw][r]{\scriptsize \shortstack[r]{FMPR\\only}}} &
        \raisebox{-.5\height}{\includegraphics[height=\imgh,keepaspectratio]{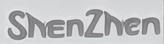}} &
        \raisebox{-.5\height}{\includegraphics[height=\imgh,keepaspectratio]{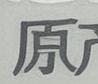}} &
        \raisebox{-.5\height}{\includegraphics[height=\imgh,keepaspectratio]{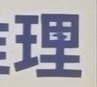}} &
        \raisebox{-.5\height}{\includegraphics[height=\imgh,keepaspectratio]{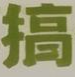}}
        \vspace{\cgap}
        \\
        
        \raisebox{-.5\height}{\makebox[\labw][r]{\scriptsize \shortstack[r]{Full\\Model}}} &
        \raisebox{-.5\height}{\includegraphics[height=\imgh,keepaspectratio]{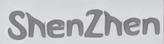}} &
        \raisebox{-.5\height}{\includegraphics[height=\imgh,keepaspectratio]{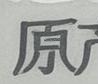}} &
        \raisebox{-.5\height}{\includegraphics[height=\imgh,keepaspectratio]{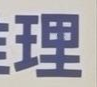}} &
        \raisebox{-.5\height}{\includegraphics[height=\imgh,keepaspectratio]{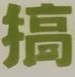}}
        
        \end{tabular}
        \vspace{-2mm}
        \caption{SURE visual details.}
        \label{fig:ablation_sure}
    \end{minipage}
    }
\vspace{-6mm}
\end{figure*}

\vspace{-1mm}
\noindent \textbf{Analysis of FMPR Euler Steps.}
We further study the effect of FMPR Euler steps $K$.
As shown in Fig.~\ref{fig:ablation_euler_steps}, increasing $K$ consistently improves perceptual realism and text fidelity, indicating that one-step rectification is insufficient for reliable prior recovery. The gains gradually saturate as more steps are used, suggesting that FMPR quickly approaches a stable region in the prior space. While $K=32$ yields slight improvements, it doubles the computational cost compared with $K=16$. We therefore choose $K=16$ as a practical trade-off between restoration performance and efficiency.

\vspace{-1mm}
\noindent \textbf{Analysis of SURE.}
We ablate SURE on RealCE-val by progressively removing its uncertainty and structural guidance. As shown in Tab.~\ref{tab:ablation_sure}, a plain residual branch without any edge information input brings limited benefit and even weakens perceptual reconstruction. Introducing deterministic edge guidance improves performance by providing complementary stroke-level information. However, deterministic guidance remains vulnerable to severe degradation, where unreliable stroke evidence may be injected under confident conditions and thus limits character-level recovery. The full model addresses this issue with uncertainty learning, allowing the structural branch to model ambiguous regions instead of enforcing a deterministic prediction. This consistently improves restoration quality and text fidelity. The zoomed-in patches in Fig.~\ref{fig:ablation_sure} further illustrate this effect. Compared with the base model and FMPR-only result, the full model produces cleaner local structures and more stable stroke topology. For the first column, it restores the letter ``e'' more accurately. And for the last three Chinese characters, the full model preserves more accurate strokes and tighter structural closure.

\vspace{-2mm}
\section{Conclusion}
\label{sec:conclusion}
\vspace{-2mm}
We proposed \textbf{PRISM}, a single-step diffusion-based framework for Text-SR that addresses two coupled ambiguities under severe degradation: unreliable text-aware prior estimation and uncertain local stroke structures. PRISM decomposes the restoration process into global prior rectification and local structure refinement. FMPR constructs a privileged prior space from paired LQ/HQ latents and learns to recover a reliable text-aware condition from degraded inputs through flow matching. SURE further injects uncertainty-aware structural residuals into the frozen restoration backbone, allowing the model to refine ambiguous stroke boundaries without over-committing to unreliable LQ edge evidence. This design preserves the efficiency of one-step diffusion restoration while improving both character fidelity and perceptual quality. Experiments on synthetic and real-world benchmarks demonstrate that PRISM achieves superior performance over representative Text-SR and text-aware restoration methods, especially under severe degradation and complex glyph structures.

\small
\bibliographystyle{unsrt}
\bibliography{arxiv_reference}

@String(IJCV = {Int. J. Comput. Vis.})

@String(CVPR= {IEEE Conf. Comput. Vis. Pattern Recog.})

@String(ICCV= {Int. Conf. Comput. Vis.})

@String(ECCV= {Eur. Conf. Comput. Vis.})

@String(TIP  = {IEEE Trans. Image Process.})

@String(ICASSP=	{ICASSP})

@String(ICLR = {Int. Conf. Learn. Represent.})

@String(IJCAI = {IJCAI})

@String(AAAI = {AAAI})

@String(CVPRW= {IEEE Conf. Comput. Vis. Pattern Recog. Worksh.})

@String(IJCV  = {IJCV})

@String(CVPR  = {CVPR})

@String(ICCV  = {ICCV})

@String(ECCV  = {ECCV})

@String(TIP   = {IEEE TIP})

@String(ICLR  = {ICLR})

@String(CVPRW= {CVPRW})

@inproceedings{zhang2021bsrgan,
  title={Designing a practical degradation model for deep blind image super-resolution},
  author={Zhang, Kai and Liang, Jingyun and Van Gool, Luc and Timofte, Radu},
  booktitle={ICCV},
  year={2021}
}

@inproceedings{wang2021realesrgan,
  title={Real-esrgan: Training real-world blind super-resolution with pure synthetic data},
  author={Wang, Xintao and Xie, Liangbin and Dong, Chao and Shan, Ying},
  booktitle={ICCV},
  year={2021}
}

@article{wang2024stablesr,
  title={Exploiting diffusion prior for real-world image super-resolution},
  author={Wang, Jianyi and Yue, Zongsheng and Zhou, Shangchen and Chan, Kelvin CK and Loy, Chen Change},
  journal={IJCV},
  year={2024}
}

@inproceedings{lin2024diffbir,
  title={Diffbir: Toward blind image restoration with generative diffusion prior},
  author={Lin, Xinqi and He, Jingwen and Chen, Ziyan and Lyu, Zhaoyang and Dai, Bo and Yu, Fanghua and Qiao, Yu and Ouyang, Wanli and Dong, Chao},
  booktitle={ECCV},
  year={2024}
}

@inproceedings{wu2024seesr,
  title={Seesr: Towards semantics-aware real-world image super-resolution},
  author={Wu, Rongyuan and Yang, Tao and Sun, Lingchen and Zhang, Zhengqiang and Li, Shuai and Zhang, Lei},
  booktitle={CVPR},
  year={2024}
}

@inproceedings{yu2024supir,
  title={Scaling up to excellence: Practicing model scaling for photo-realistic image restoration in the wild},
  author={Yu, Fanghua and Gu, Jinjin and Li, Zheyuan and Hu, Jinfan and Kong, Xiangtao and Wang, Xintao and He, Jingwen and Qiao, Yu and Dong, Chao},
  booktitle={CVPR},
  year={2024}
}

@inproceedings{wang2024sinsr,
  title={Sinsr: diffusion-based image super-resolution in a single step},
  author={Wang, Yufei and Yang, Wenhan and Chen, Xinyuan and Wang, Yaohui and Guo, Lanqing and Chau, Lap-Pui and Liu, Ziwei and Qiao, Yu and Kot, Alex C and Wen, Bihan},
  booktitle={CVPR},
  year={2024}
}

@inproceedings{wu2024osediff,
  title={One-step effective diffusion network for real-world image super-resolution},
  author={Wu, Rongyuan and Sun, Lingchen and Ma, Zhiyuan and Zhang, Lei},
  booktitle={NeurIPS},
  year={2024}
}

@inproceedings{yue2025invsr,
  title={Arbitrary-steps image super-resolution via diffusion inversion},
  author={Yue, Zongsheng and Liao, Kang and Loy, Chen Change},
  booktitle={CVPR},
  year={2025}
}

@article{lin2025hypir,
  title={Harnessing diffusion-yielded score priors for image restoration},
  author={Lin, Xinqi and Yu, Fanghua and Hu, Jinfan and You, Zhiyuan and Shi, Wu and Ren, Jimmy S and Gu, Jinjin and Dong, Chao},
  journal={SIGGRAPH Asia},
  year={2025}
}

@inproceedings{ldm,
  title={High-Resolution Image Synthesis With Latent Diffusion Models},
  author={Rombach, Robin and Blattmann, Andreas and Lorenz, Dominik and Esser, Patrick and Ommer, Bj\"orn},
  booktitle={CVPR},
  year={2022},
}

@article{podell2023sdxl,
  title={Sdxl: Improving latent diffusion models for high-resolution image synthesis},
  author={Podell, Dustin and English, Zion and Lacey, Kyle and Blattmann, Andreas and Dockhorn, Tim and M{\"u}ller, Jonas and Penna, Joe and Rombach, Robin},
  journal={arXiv preprint arXiv:2307.01952},
  year={2023}
}

@inproceedings{peebles2023dit,
  title={Scalable diffusion models with transformers},
  author={Peebles, William and Xie, Saining},
  booktitle={ICCV},
  year={2023}
}

@inproceedings{esser2024sd3,
  title={Scaling rectified flow transformers for high-resolution image synthesis},
  author={Esser, Patrick and Kulal, Sumith and Blattmann, Andreas and Entezari, Rahim and M{\"u}ller, Jonas and Saini, Harry and Levi, Yam and Lorenz, Dominik and Sauer, Axel and Boesel, Frederic and others},
  booktitle={ICML},
  year={2024}
}

@inproceedings{dong2025tsdsr,
  title={Tsd-sr: One-step diffusion with target score distillation for real-world image super-resolution},
  author={Dong, Linwei and Fan, Qingnan and Guo, Yihong and Wang, Zhonghao and Zhang, Qi and Chen, Jinwei and Luo, Yawei and Zou, Changqing},
  booktitle={CVPR},
  year={2025}
}

@inproceedings{duan2025dit4sr,
  title={Dit4sr: Taming diffusion transformer for real-world image super-resolution},
  author={Duan, Zheng-Peng and Zhang, Jiawei and Jin, Xin and Zhang, Ziheng and Xiong, Zheng and Zou, Dongqing and Ren, Jimmy S and Guo, Chunle and Li, Chongyi},
  booktitle={ICCV},
  year={2025}
}

@misc{flux2024,
    author={Black Forest Labs},
    title={FLUX},
    year={2024},
    howpublished={\url{https://github.com/black-forest-labs/flux}},
}

@inproceedings{li2025fluxsr,
  title={One diffusion step to real-world super-resolution via flow trajectory distillation},
  author={Li, Jianze and Cao, Jiezhang and Guo, Yong and Li, Wenbo and Zhang, Yulun},
  booktitle={ICML},
  year={2025}
}

@inproceedings{wang2020tsrn,
  title={Scene text image super-resolution in the wild},
  author={Wang, Wenjia and Xie, Enze and Liu, Xuebo and Wang, Wenhai and Liang, Ding and Shen, Chunhua and Bai, Xiang},
  booktitle={ECCV},
  year={2020},
}

@inproceedings{chen2021tbsrn,
  title={Scene text telescope: Text-focused scene image super-resolution},
  author={Chen, Jingye and Li, Bin and Xue, Xiangyang},
  booktitle={CVPR},
  year={2021}
}

@inproceedings{ma2022tatt,
  title={A text attention network for spatial deformation robust scene text image super-resolution},
  author={Ma, Jianqi and Liang, Zhetong and Zhang, Lei},
  booktitle={CVPR},
  year={2022}
}

@inproceedings{li2023marconet,
  title={Learning generative structure prior for blind text image super-resolution},
  author={Li, Xiaoming and Zuo, Wangmeng and Loy, Chen Change},
  booktitle={CVPR},
  year={2023}
}

@article{ma2023tpgsr,
  title={Text prior guided scene text image super-resolution},
  author={Ma, Jianqi and Guo, Shi and Zhang, Lei},
  journal={IEEE TIP},
  year={2023}
}

@inproceedings{guo2023lemma,
  title={Towards robust scene text image super-resolution via explicit location enhancement},
  author={Guo, Hang and Dai, Tao and Meng, Guanghao and Xia, Shu-Tao},
  booktitle={IJCAI},
  year={2023}
}

@inproceedings{zhao2023stirer,
  title={STIRER: A unified model for low-resolution scene text image recovery and recognition},
  author={Zhao, Minyi and Xuyang, Shijie and Guan, Jihong and Zhou, Shuigeng},
  booktitle={ACM MM},
  year={2023}
}

@inproceedings{zhu2023tsan,
  title={Gradient-based graph attention for scene text image super-resolution},
  author={Zhu, Xiangyuan and Guo, Kehua and Fang, Hui and Ding, Rui and Wu, Zheng and Schaefer, Gerald},
  booktitle={AAAI},
  year={2023}
}

@inproceedings{zhu2023dpmn,
  title={Improving scene text image super-resolution via dual prior modulation network},
  author={Zhu, Shipeng and Zhao, Zuoyan and Fang, Pengfei and Xue, Hui},
  booktitle={AAAI},
  year={2023}
}

@inproceedings{singh2024dcdm,
  title={Dcdm: Diffusion-conditioned-diffusion model for scene text image super-resolution},
  author={Singh, Shrey and Keserwani, Prateek and Iwamura, Masakazu and Roy, Partha Pratim},
  booktitle={ECCV},
  year={2024}
}

@inproceedings{zhao2024pean,
  title={Pean: A diffusion-based prior-enhanced attention network for scene text image super-resolution},
  author={Zhao, Zuoyan and Xue, Hui and Fang, Pengfei and Zhu, Shipeng},
  booktitle={ACM MM},
  year={2024}
}

@inproceedings{zhang2024difftsr,
  title={Diffusion-based blind text image super-resolution},
  author={Zhang, Yuzhe and Zhang, Jiawei and Li, Hao and Wang, Zhouxia and Hou, Luwei and Zou, Dongqing and Bian, Liheng},
  booktitle={CVPR},
  year={2024}
}

@inproceedings{yuan2025stylesrn,
  title={StyleSRN: Scene Text Image Super-Resolution with Text Style Embedding},
  author={Yuan, Shengrong and Wang, Runmin and Hao, Ke and Ma, Xuqi and Gao, Changxin and Liu, Li and Sang, Nong},
  booktitle={ICCV},
  year={2025}
}

@article{li2025marconetplusplus,
  title={Enhanced Generative Structure Prior for Chinese Text Image Super-Resolution},
  author={Li, Xiaoming and Zuo, Wangmeng and Loy, Chen Change},
  journal={IEEE TPAMI},
  year={2025}
}

@inproceedings{wei2025glyphsr,
  title={GlyphSR: A Simple Glyph-Aware Framework for Scene Text Image Super-Resolution},
  author={Wei, Baole and Zhou, Yuxuan and Gao, Liangcai and Tang, Zhi},
  booktitle={AAAI},
  year={2025}
}

@inproceedings{hu2025tadisr,
  title={Text-Aware Real-World Image Super-Resolution via Diffusion Model with Joint Segmentation Decoders},
  author={Hu, Qiming and Fan, Linlong and Luo, Yiyan and Yu, Yuhang and Guo, Xiaojie and Fan, Qingnan},
  booktitle={NeurIPS},
  year={2025}
}

@inproceedings{min2026terediff,
  title={Text-Aware Image Restoration with Diffusion Models},
  author={Min, Jaewon and Kim, Jin Hyeon and Cho, Paul Hyunbin and Lee, Jaeeun and Park, Jihye and Park, Minkyu and Kim, Sangpil and Park, Hyunhee and Kim, Seungryong},
  booktitle={ICLR},
  year={2026}
}

@inproceedings{he2026textsdiff,
  title={TEXTS-Diff: TEXTS-Aware Diffusion Model for Real-World Text Image Super-Resolution},
  author={He, Haodong and Zhan, Xin and Bai, Yancheng and Lan, Rui and Sun, Lei and Chu, Xiangxiang},
  booktitle={ICASSP},
  year={2026}
}

@article{vapnik2015learning,
  title={Learning using privileged information: similarity control and knowledge transfer},
  author={Vapnik, Vladimir and Izmailov, Rauf},
  journal={JMLR},
  year={2015},
}

@inproceedings{lee2020pisr,
  title={Learning with privileged information for efficient image super-resolution},
  author={Lee, Wonkyung and Lee, Junghyup and Kim, Dohyung and Ham, Bumsub},
  booktitle={ECCV},
  year={2020},
}

@inproceedings{xia2023diffir,
  title={Diffir: Efficient diffusion model for image restoration},
  author={Xia, Bin and Zhang, Yulun and Wang, Shiyin and Wang, Yitong and Wu, Xinglong and Tian, Yapeng and Yang, Wenming and Van Gool, Luc},
  booktitle={ICCV},
  year={2023}
}

@inproceedings{lipman2023flow,
  title={Flow matching for generative modeling},
  author={Lipman, Yaron and Chen, Ricky TQ and Ben-Hamu, Heli and Nickel, Maximilian and Le, Matt},
  booktitle={ICLR},
  year={2023}
}

@inproceedings{liu2022flow,
  title={Flow straight and fast: Learning to generate and transfer data with rectified flow},
  author={Liu, Xingchao and Gong, Chengyue and Liu, Qiang},
  booktitle={ICLR},
  year={2023}
}

@inproceedings{kendall2017uncertainties,
  title={What uncertainties do we need in bayesian deep learning for computer vision?},
  author={Kendall, Alex and Gal, Yarin},
  booktitle={NeurIPS},
  year={2017}
}

@article{yu2021ctr,
  title={Benchmarking chinese text recognition: Datasets, baselines, and an empirical study},
  author={Yu, Haiyang and Chen, Jingye and Li, Bin and Ma, Jianqi and Guan, Mengnan and Xu, Xixi and Wang, Xiaocong and Qu, Shaobo and Xue, Xiangyang},
  journal={arXiv preprint arXiv:2112.15093},
  year={2021}
}

@inproceedings{ma2023realce,
  title={A benchmark for chinese-english scene text image super-resolution},
  author={Ma, Jianqi and Liang, Zhetong and Xiang, Wangmeng and Yang, Xi and Zhang, Lei},
  booktitle={ICCV},
  year={2023}
}

@inproceedings{fang2023UFPNet,
  author={Fang, Zhenxuan and Wu, Fangfang and Dong, Weisheng and Li, Xin and Wu, Jinjian and Shi, Guangming},
  title={Self-supervised Non-uniform Kernel Estimation with Flow-based Motion Prior for Blind Image Deblurring},
  booktitle={ CVPR},
  year={2023}
}

@inproceedings{NEURIPS2021_88a19961,
  title={Uncertainty-Driven Loss for Single Image Super-Resolution},
  author={Ning, Qian and Dong, Weisheng and Li, Xin and Wu, Jinjian and Shi, GUANGMING},
  booktitle={NeurIPS},
  year={2021}
}

@inproceedings{hu2022lora,
  title={Lora: Low-rank adaptation of large language models.},
  author={Hu, Edward J and Shen, Yelong and Wallis, Phillip and Allen-Zhu, Zeyuan and Li, Yuanzhi and Wang, Shean and Wang, Liang and Chen, Weizhu and others},
  booktitle={ICLR},
  year={2022}
}

@inproceedings{zhang2018lpips,
  title={The Unreasonable Effectiveness of Deep Features as a Perceptual Metric},
  author={Zhang, Richard and Isola, Phillip and Efros, Alexei A and Shechtman, Eli and Wang, Oliver},
  booktitle={CVPR},
  year={2018}
}

@inproceedings{heusel2017fid,
  title={Gans trained by a two time-scale update rule converge to a local nash equilibrium},
  author={Heusel, Martin and Ramsauer, Hubert and Unterthiner, Thomas and Nessler, Bernhard and Hochreiter, Sepp},
  booktitle={NeurIPS},
  year={2017}
}

@inproceedings{ke2021musiq,
  title={Musiq: Multi-scale image quality transformer},
  author={Ke, Junjie and Wang, Qifei and Wang, Yilin and Milanfar, Peyman and Yang, Feng},
  booktitle={ICCV},
  year={2021}
}

@inproceedings{yang2022maniqa,
  title={Maniqa: Multi-dimension attention network for no-reference image quality assessment},
  author={Yang, Sidi and Wu, Tianhe and Shi, Shuwei and Lao, Shanshan and Gong, Yuan and Cao, Mingdeng and Wang, Jiahao and Yang, Yujiu},
  booktitle={CVPRW},
  year={2022}
}

@inproceedings{wang2023clipiqa,
  title={Exploring clip for assessing the look and feel of images},
  author={Wang, Jianyi and Chan, Kelvin CK and Loy, Chen Change},
  booktitle={AAAI},
  year={2023}
}

@inproceedings{ddpm,
  author={Ho, Jonathan and Jain, Ajay and Abbeel, Pieter},
  title={Denoising diffusion probabilistic models},
  year={2020},
  booktitle={NeurIPS},
}

@inproceedings{li2024distillation,
  title={Distillation-Free One-Step Diffusion for Real-World Image Super-Resolution},
  author={Li, Jianze and Cao, Jiezhang and Zou, Zichen and Su, Xiongfei and Yuan, Xin and Zhang, Yulun and Guo, Yong and Yang, Xiaokang},
  booktitle={NeurIPS},
  year={2025}
}

@article{cui2025paddleocr,
  title={Paddleocr 3.0 technical report},
  author={Cui, Cheng and Sun, Ting and Lin, Manhui and Gao, Tingquan and Zhang, Yubo and Liu, Jiaxuan and Wang, Xueqing and Zhang, Zelun and Zhou, Changda and Liu, Hongen and others},
  journal={arXiv preprint arXiv:2507.05595},
  year={2025}
}

@inproceedings{wang2025osdface,
  title={{OSDFace}: One-Step Diffusion Model for Face Restoration},
  author={Wang, Jingkai and Gong, Jue and Zhang, Lin and Chen, Zheng and Liu, Xing and Gu, Hong and Liu, Yutong and Zhang, Yulun and Yang, Xiaokang},
  booktitle={CVPR},
  year={2025},
}

@article{liu2025osdd,
  title={One-Step Diffusion Model for Image Motion-Deblurring}, 
  author={Liu, Xiaoyang and Wang, Yuquan and Chen, Zheng and Cao, Jiezhang and Zhang, He and Zhang, Yulun and Yang, Xiaokang},
  journal={arXiv preprint arXiv:2503.06537},
  year={2025}
}

@inproceedings{liu2025fidediff,
  title={FideDiff: Efficient Diffusion Model for High-Fidelity Image Motion Deblurring}, 
  author={Liu, Xiaoyang and Zhou, Zhengyan and Xu, Zihang and Cao, Jiezhang and Chen, Zheng and Zhang, Yulun},
  booktitle={ICLR},
  year={2025}
}

\newpage
\appendix

\section{Details of BTL Dataset Construction}
\label{app:btl_dataset}
\paragraph{Motivation and source data.}
Existing text-image datasets cover different aspects of Text-SR, but no single resource fully matches our training setting of high-quality Chinese-English text-line super-resolution. TextZoom~\citep{wang2020tsrn} provides real paired LR-HR scene text images and has been widely used as a benchmark for scene Text-SR, but it mainly focuses on English text and often suffers from varying image quality. RealCE~\citep{ma2023realce} further introduces a Chinese-English scene Text-SR benchmark with an emphasis on structurally complex Chinese characters, but its scale is relatively limited.

For constructing BTL, we use two annotated text-image sources as real-image candidate pools for HQ text-line crops. The first is CTR~\citep{yu2021ctr}, a large-scale Chinese text recognition benchmark built from multiple scene-text datasets. Although CTR is designed for recognition rather than super-resolution, it provides a large number of cropped text-line images with transcripts and contains a high proportion of Chinese samples, making it suitable for selecting Chinese real-text candidates. The second is SA-Text~\citep{min2026terediff}, a large-scale text-aware image restoration dataset built from high-quality scene images with detailed text annotations. Since SA-Text provides abundant English text instances and high-quality visual content, we use it as the main source for English real-text candidates. In addition, we generate synthetic HQ text-line images following the rendering strategy of MARCONet~\citep{li2023marconet}. The final BTL dataset combines curated real HQ crops from CTR and SA-Text with synthetic HQ text-line images, balancing real-world appearance, language coverage, and controllable text-line diversity.

\paragraph{Language statistics of source pools.}
We analyze the transcript distribution of the two real-image source pools used for BTL construction, i.e., CTR~\citep{yu2021ctr} and SA-Text~\citep{min2026terediff}. We categorize each candidate according to its transcript. \emph{Chinese} denotes samples containing at least one Chinese character; \emph{English} denotes samples containing English letters but no Chinese characters; \emph{Digit} denotes samples containing only digits after removing punctuation; and the remaining samples are grouped as \emph{Others}. As shown in Tab.~\ref{tab:source_language_stats}, CTR contains a substantially larger proportion of Chinese text, while SA-Text provides more English candidates. This supports our source allocation strategy: Chinese candidates are selected from CTR, whereas English candidates are selected from SA-Text.

\paragraph{Real HQ text-line crop selection.}
All candidate crops from CTR~\citep{yu2021ctr} and SA-Text~\citep{min2026terediff} are filtered using the following protocol. First, each crop is resized to a fixed height of 128 using bicubic interpolation while preserving its aspect ratio, so that all candidates are assessed under a consistent resolution. We then retain samples whose aspect ratio falls between 2 and 8 and whose transcript length is no more than 24 characters. These constraints remove unsuitable text-line geometries, such as extremely short, overly long, or densely annotated instances, and match our target setting of high-quality bilingual text-line SR. Finally, we rank the retained candidates using no-reference IQA metrics to select visually reliable crops. Within each language/source group, MUSIQ, MANIQA, and CLIP-IQA scores are converted into percentile ranks in $[0,1]$, denoted as $\mathcal{R}_{\mathrm{MUSIQ}}$, $\mathcal{R}_{\mathrm{MANIQA}}$, and $\mathcal{R}_{\mathrm{CLIP-IQA}}$, respectively. The final quality score is computed as
\[
\mathcal{Q} = 0.50 \mathcal{R}_{\mathrm{MUSIQ}} + 0.35 \mathcal{R}_{\mathrm{MANIQA}} + 0.15 \mathcal{R}_{\mathrm{CLIP-IQA}}.
\]
We sort candidates by $\mathcal{Q}$ within each group and allocate the selection quota according to the group proportion in the retained candidate pool, preserving the original group distribution in the curated subset. The resulting curated real HQ subset contains 50K images, including 31,567 Chinese samples and 1,084 digit-only samples from CTR, and 16,634 English samples and 715 digit-only samples from SA-Text.

\paragraph{Synthetic HQ text-line images.}
To increase scale and content diversity, we additionally generate 50K synthetic HQ text-line images following the rendering strategy of MARCONet~\citep{li2023marconet}. Rendered samples provide controllable high-quality text-line images with diverse transcripts, fonts, and layouts, while the curated real subset contributes real image statistics and background-text interactions. Combining these two sources allows BTL to balance controllability and real-world appearance.

\paragraph{LQ synthesis and dataset split.}
For each HQ text-line image, we synthesize the corresponding LQ input using degradation pipelines based on BSRGAN~\citep{zhang2021bsrgan} and Real-ESRGAN~\citep{wang2021realesrgan}. The final BTL dataset contains 100K HQ images, consisting of 50K curated real crops and 50K rendered text-line images. We preserve the real/synthetic ratio and split BTL into 80K training samples and 20K testing samples, denoted as BTL-train and BTL-test, respectively. Tab.~\ref{tab:btl_composition} summarizes the final dataset composition.

\paragraph{Effect of data composition.}
We further examine the effect of different HQ data sources by training PRISM with three data configurations and evaluating them on RealCE-val. The first configuration, denoted as Synth-train, uses synthetic text lines generated following the rendering strategy of MARCONet. The second configuration, denoted as CTR-train, uses real text crops from the training split of CTR. The third configuration is the proposed BTL-train, which combines curated real crops and rendered text lines. All three PRISM variants are trained with the same training configuration and evaluated on the same RealCE-val set. As shown in Fig.~\ref{fig:app_dataset_compare}, PRISM trained only on Synth-train tends to produce sharp but less naturally integrated strokes in some real-world cases, where the restored text may appear separated from the background. PRISM trained only on CTR-train produces more conservative results, but the restored text is often less clear. In comparison, PRISM trained on BTL-train provides a better balance between text sharpness and real-world appearance. This qualitative analysis suggests that combining curated real crops and rendered text lines is beneficial for real-world Text-SR.

\begin{table}[t]
\centering
\setlength{\tabcolsep}{6pt}
\begin{tabular}{l|l|ccccc}
\toprule
Source & Total & Chinese & English & Digit & Others \\
\midrule
CTR     & 636,455 & 519,974 & 82,484 & 31,188 & 2,809 \\
SA-Text & 271,126 & 1,743 & 236,258 & 29,336 & 3,789 \\
\bottomrule
\end{tabular}
\vspace{4mm}%
\caption{Language statistics of the real-image source pools used for BTL construction.}
\label{tab:source_language_stats}
\end{table}
\begin{table}[t]
\centering
\setlength{\tabcolsep}{6pt}
\begin{tabular}{l|ccc}
\toprule
Subset & Curated Real HQ & Synthetic HQ & Total \\
\midrule
BTL-train & 40,000 & 40,000 & 80,000 \\
BTL-test  & 10,000 & 10,000 & 20,000 \\
\midrule
Total     & 50,000 & 50,000 & 100,000 \\
\bottomrule
\end{tabular}
\vspace{4mm}%
\caption{Final composition of BTL. For each HQ image, the corresponding LQ input is synthesized using the degradation pipelines described in Sec.~\ref{app:btl_dataset}.}
\label{tab:btl_composition}
\end{table}
\begin{figure*}[t]
\centering

\setlength{\tabcolsep}{0pt}
\renewcommand{\arraystretch}{0}

\hspace{-15mm}%
\begin{adjustbox}{max width=1.1\textwidth}
\begin{tabular}{@{}r@{\hspace{2mm}}c@{\hspace{1mm}}c@{\hspace{1mm}}c@{\hspace{1mm}}c@{\hspace{1mm}}c@{}}

\raisebox{-.5\height}{\makebox[0.2\textwidth][r]{\large GT}} &
\raisebox{-.5\height}{\includegraphics[height=0.1\textwidth,keepaspectratio]{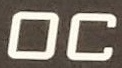}} &
\raisebox{-.5\height}{\includegraphics[height=0.1\textwidth,keepaspectratio]{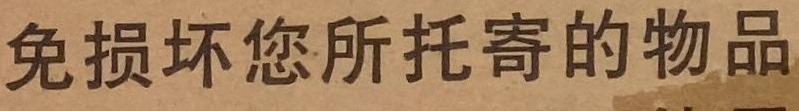}} &
\raisebox{-.5\height}{\includegraphics[height=0.1\textwidth,keepaspectratio]{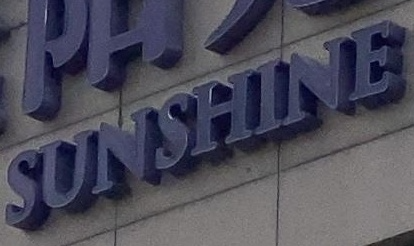}} &
\raisebox{-.5\height}{\includegraphics[height=0.1\textwidth,keepaspectratio]{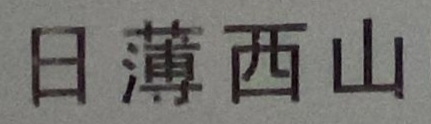}}
\vspace{1mm}%
\\[2pt]

\raisebox{-.5\height}{\makebox[0.2\textwidth][r]{\large LR}} &
\raisebox{-.5\height}{\includegraphics[height=0.1\textwidth,keepaspectratio]{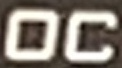}} &
\raisebox{-.5\height}{\includegraphics[height=0.1\textwidth,keepaspectratio]{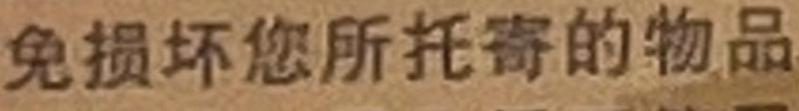}} &
\raisebox{-.5\height}{\includegraphics[height=0.1\textwidth,keepaspectratio]{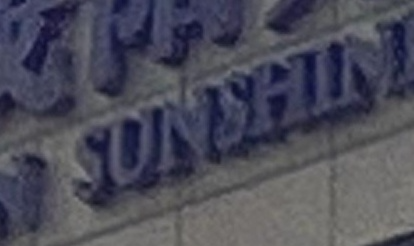}} &
\raisebox{-.5\height}{\includegraphics[height=0.1\textwidth,keepaspectratio]{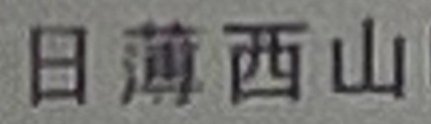}}
\vspace{1mm}%
\\[2pt]

\raisebox{-.5\height}{\makebox[0.2\textwidth][r]{\large Synth-train}} &
\raisebox{-.5\height}{\includegraphics[height=0.1\textwidth,keepaspectratio]{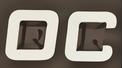}} &
\raisebox{-.5\height}{\includegraphics[height=0.1\textwidth,keepaspectratio]{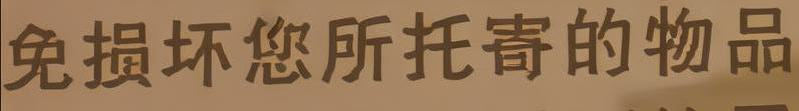}} &
\raisebox{-.5\height}{\includegraphics[height=0.1\textwidth,keepaspectratio]{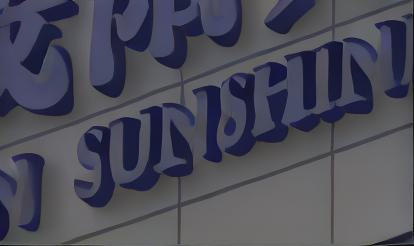}} &
\raisebox{-.5\height}{\includegraphics[height=0.1\textwidth,keepaspectratio]{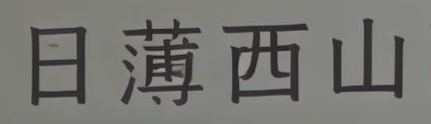}}
\vspace{1mm}%
\\[2pt]

\raisebox{-.5\height}{\makebox[0.2\textwidth][r]{\large CTR-train}} &
\raisebox{-.5\height}{\includegraphics[height=0.1\textwidth,keepaspectratio]{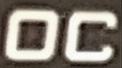}} &
\raisebox{-.5\height}{\includegraphics[height=0.1\textwidth,keepaspectratio]{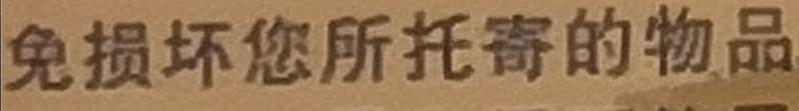}} &
\raisebox{-.5\height}{\includegraphics[height=0.1\textwidth,keepaspectratio]{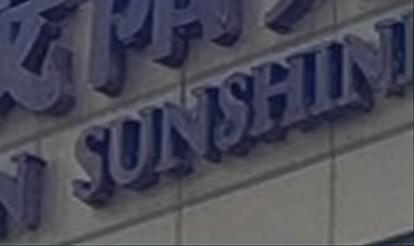}} &
\raisebox{-.5\height}{\includegraphics[height=0.1\textwidth,keepaspectratio]{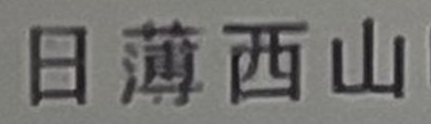}}
\vspace{1mm}%
\\[2pt]

\raisebox{-.5\height}{\makebox[0.2\textwidth][r]{\large BTL-train}} &
\raisebox{-.5\height}{\includegraphics[height=0.1\textwidth,keepaspectratio]{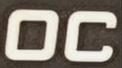}} &
\raisebox{-.5\height}{\includegraphics[height=0.1\textwidth,keepaspectratio]{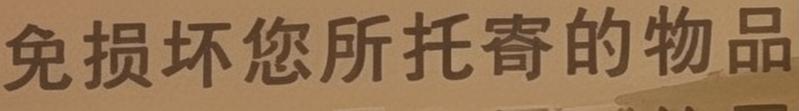}} &
\raisebox{-.5\height}{\includegraphics[height=0.1\textwidth,keepaspectratio]{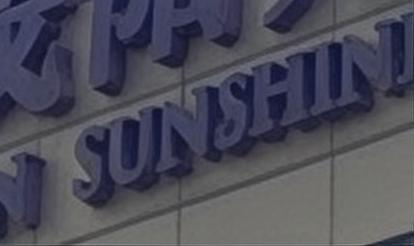}} &
\raisebox{-.5\height}{\includegraphics[height=0.1\textwidth,keepaspectratio]{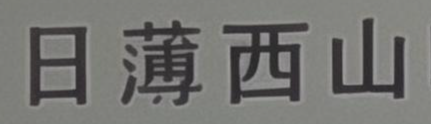}}

\end{tabular}
\end{adjustbox}

\vspace{5mm}%
\hspace{-15mm}%
\begin{adjustbox}{max width=1.1\textwidth}
\begin{tabular}{@{}r@{\hspace{2mm}}c@{\hspace{1mm}}c@{\hspace{1mm}}c@{\hspace{1mm}}c@{\hspace{1mm}}c@{}}

\raisebox{-.5\height}{\makebox[0.2\textwidth][r]{\large GT}} &
\raisebox{-.5\height}{\includegraphics[height=0.1\textwidth,keepaspectratio]{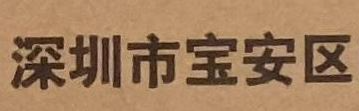}} &
\raisebox{-.5\height}{\includegraphics[height=0.1\textwidth,keepaspectratio]{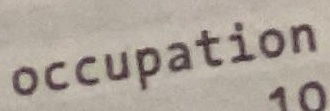}} &
\raisebox{-.5\height}{\includegraphics[height=0.1\textwidth,keepaspectratio]{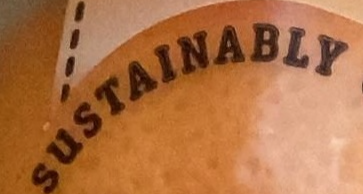}} &
\raisebox{-.5\height}{\includegraphics[height=0.1\textwidth,keepaspectratio]{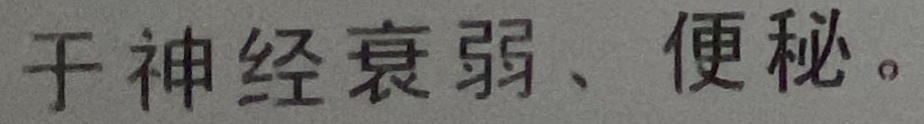}}
\vspace{1mm}%
\\[2pt]

\raisebox{-.5\height}{\makebox[0.2\textwidth][r]{\large LR}} &
\raisebox{-.5\height}{\includegraphics[height=0.1\textwidth,keepaspectratio]{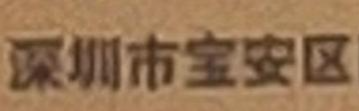}} &
\raisebox{-.5\height}{\includegraphics[height=0.1\textwidth,keepaspectratio]{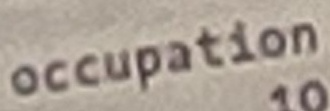}} &
\raisebox{-.5\height}{\includegraphics[height=0.1\textwidth,keepaspectratio]{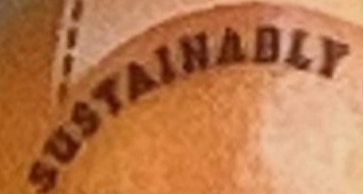}} &
\raisebox{-.5\height}{\includegraphics[height=0.1\textwidth,keepaspectratio]{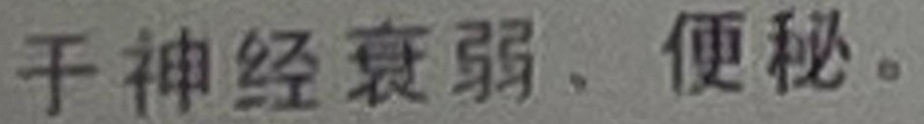}}
\vspace{1mm}%
\\[2pt]

\raisebox{-.5\height}{\makebox[0.2\textwidth][r]{\large Synth-train}} &
\raisebox{-.5\height}{\includegraphics[height=0.1\textwidth,keepaspectratio]{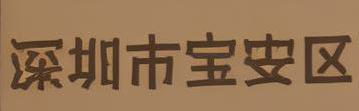}} &
\raisebox{-.5\height}{\includegraphics[height=0.1\textwidth,keepaspectratio]{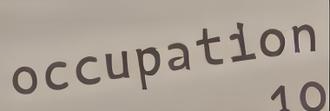}} &
\raisebox{-.5\height}{\includegraphics[height=0.1\textwidth,keepaspectratio]{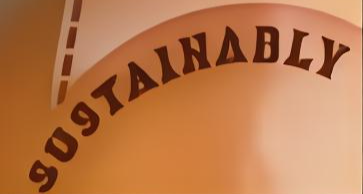}} &
\raisebox{-.5\height}{\includegraphics[height=0.1\textwidth,keepaspectratio]{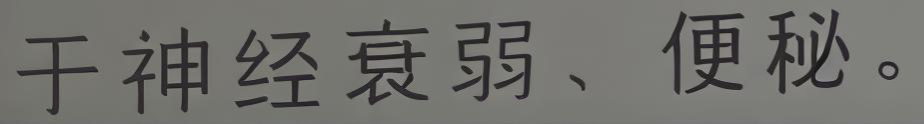}}
\vspace{1mm}%
\\[2pt]

\raisebox{-.5\height}{\makebox[0.2\textwidth][r]{\large CTR-train}} &
\raisebox{-.5\height}{\includegraphics[height=0.1\textwidth,keepaspectratio]{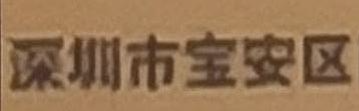}} &
\raisebox{-.5\height}{\includegraphics[height=0.1\textwidth,keepaspectratio]{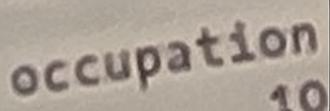}} &
\raisebox{-.5\height}{\includegraphics[height=0.1\textwidth,keepaspectratio]{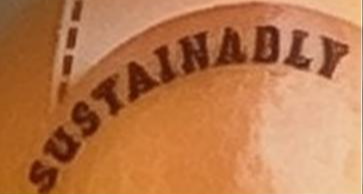}} &
\raisebox{-.5\height}{\includegraphics[height=0.1\textwidth,keepaspectratio]{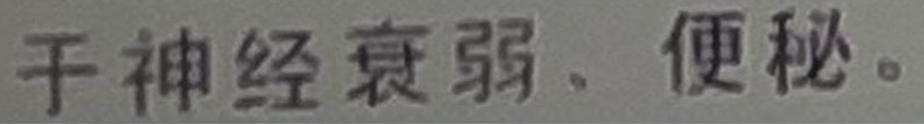}}
\vspace{1mm}%
\\[2pt]

\raisebox{-.5\height}{\makebox[0.2\textwidth][r]{\large BTL-train}} &
\raisebox{-.5\height}{\includegraphics[height=0.1\textwidth,keepaspectratio]{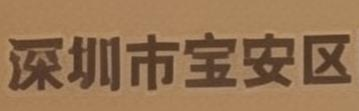}} &
\raisebox{-.5\height}{\includegraphics[height=0.1\textwidth,keepaspectratio]{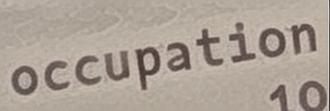}} &
\raisebox{-.5\height}{\includegraphics[height=0.1\textwidth,keepaspectratio]{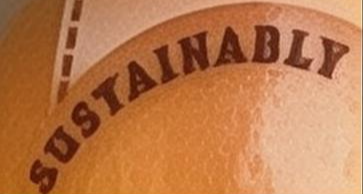}} &
\raisebox{-.5\height}{\includegraphics[height=0.1\textwidth,keepaspectratio]{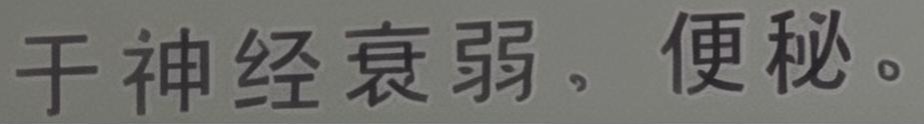}}

\end{tabular}
\end{adjustbox}

\caption{Effect of training data composition on RealCE-val for $\times 4$ super-resolution. We compare PRISM trained with three data configurations: Synth-train, CTR-train, and BTL-train.}
\label{fig:app_dataset_compare}
\end{figure*}

\section{Inference Speed Analysis}
Table~\ref{tab:speed} presents a detailed inference speed comparison on RealCE-val among different methods under the $\times4$ setting, including the number of sampling steps and per-image inference time. For a fair comparison, we select images from RealCE-val whose heights range from 50 to 100 pixels and whose widths range from 300 to 500 pixels and then resize them to the target resolution. For non-diffusion-based methods, we measure runtime with a fixed LR input size of $32 \times 128$ and a fixed HR output size of $128 \times 512$. For diffusion-based methods, both the input and output are fixed to $128 \times 512$. All methods are evaluated with batch size 1 on a single RTX 4090, and the measured runtime includes the full inference pipeline except file I/O. As shown in Tab.~\ref{tab:speed}, PRISM achieves substantially faster inference than existing diffusion-based Text-SR methods. Compared with DiffTSR~\citep{zhang2024difftsr} which uses 200 sampling steps, and TeReDiff~\citep{min2026terediff} which uses 50 steps, PRISM requires only one step and takes around 80 milliseconds per image. Notably, its runtime is comparable to the non-diffusion-based Text-SR methods. These results show that PRISM retains restoration capability while achieving practical one-step inference efficiency.
\begin{table}[t]
\centering
\begin{adjustbox}{max width=\textwidth, center}
\begin{tabular}{l|cccc|ccc}
\toprule
 & TSRN & TBSRN & TATT & MARCONet & DiffTSR-s200 & TeReDiff-s50 & $\mathbf{Ours}$ \\
\midrule
Inference Time / ms  & 6.14 & 10.48 & 25.94 & 320.11 & 10702.98 & 5273.42 & \textbf{84.74} \\
\bottomrule
\end{tabular}
\end{adjustbox}
\vspace{2mm}%
\caption{Inference speed comparison ($\times 4$) on RealCE-val among different methods. Diffusion-based methods use different numbers of inference steps: DiffTSR~\citep{zhang2024difftsr} uses 200 steps, TeReDiff~\citep{min2026terediff} uses 50 steps, and ours uses only one step.}
\label{tab:speed}
\end{table}

\section{More Visualizations}
\begin{figure*}[t]
\centering

\setlength{\tabcolsep}{0pt}
\renewcommand{\arraystretch}{0}

\hspace{-15mm}%
\begin{adjustbox}{max width=1.1\textwidth}
\begin{tabular}{@{}r@{\hspace{2mm}}c@{\hspace{1mm}}c@{\hspace{1mm}}c@{\hspace{1mm}}c@{\hspace{1mm}}c@{}}

\raisebox{-.5\height}{\makebox[0.2\textwidth][r]{\large GT}} &
\raisebox{-.5\height}{\includegraphics[height=0.1\textwidth,keepaspectratio]{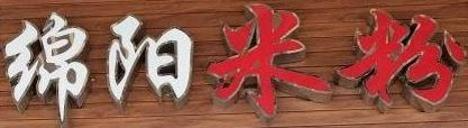}} &
\raisebox{-.5\height}{\includegraphics[height=0.1\textwidth,keepaspectratio]{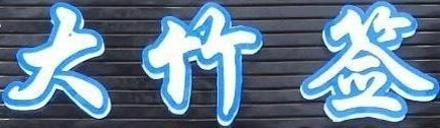}} &
\raisebox{-.5\height}{\includegraphics[height=0.1\textwidth,keepaspectratio]{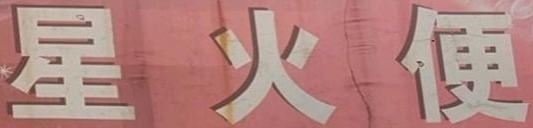}} &
\raisebox{-.5\height}{\includegraphics[height=0.1\textwidth,keepaspectratio]{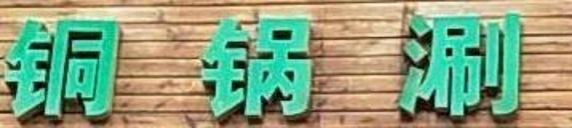}} &
\raisebox{-.5\height}{\includegraphics[height=0.1\textwidth,keepaspectratio]{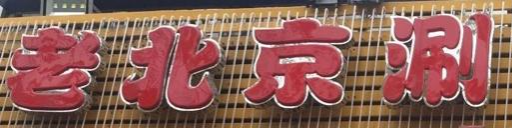}}
\vspace{1mm}
\\

\raisebox{-.5\height}{\makebox[0.2\textwidth][r]{\large LR}} &
\raisebox{-.5\height}{\includegraphics[height=0.1\textwidth,keepaspectratio]{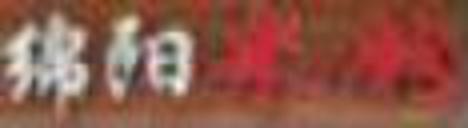}} &
\raisebox{-.5\height}{\includegraphics[height=0.1\textwidth,keepaspectratio]{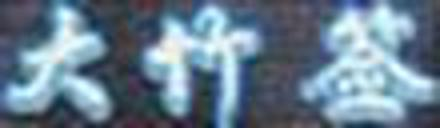}} &
\raisebox{-.5\height}{\includegraphics[height=0.1\textwidth,keepaspectratio]{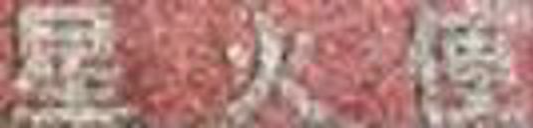}} &
\raisebox{-.5\height}{\includegraphics[height=0.1\textwidth,keepaspectratio]{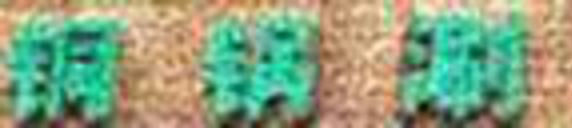}} &
\raisebox{-.5\height}{\includegraphics[height=0.1\textwidth,keepaspectratio]{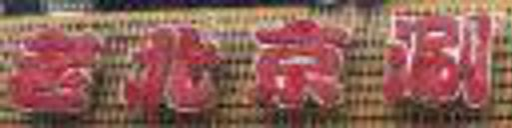}}
\vspace{1mm}
\\

\raisebox{-.5\height}{\makebox[0.2\textwidth][r]{\large TSRN}} &
\raisebox{-.5\height}{\includegraphics[height=0.1\textwidth,keepaspectratio]{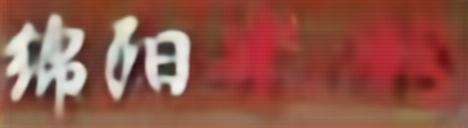}} &
\raisebox{-.5\height}{\includegraphics[height=0.1\textwidth,keepaspectratio]{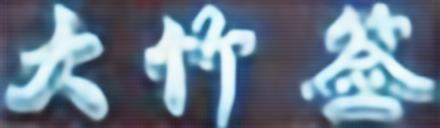}} &
\raisebox{-.5\height}{\includegraphics[height=0.1\textwidth,keepaspectratio]{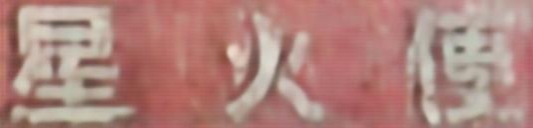}} &
\raisebox{-.5\height}{\includegraphics[height=0.1\textwidth,keepaspectratio]{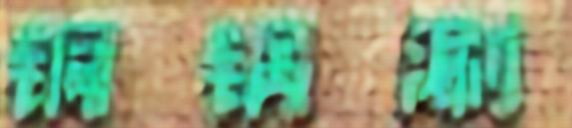}} &
\raisebox{-.5\height}{\includegraphics[height=0.1\textwidth,keepaspectratio]{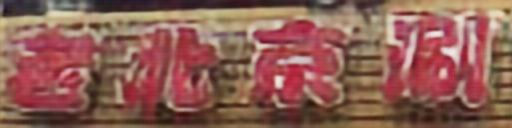}}
\vspace{1mm}
\\

\raisebox{-.5\height}{\makebox[0.2\textwidth][r]{\large TBSRN}} &
\raisebox{-.5\height}{\includegraphics[height=0.1\textwidth,keepaspectratio]{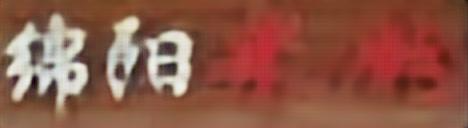}} &
\raisebox{-.5\height}{\includegraphics[height=0.1\textwidth,keepaspectratio]{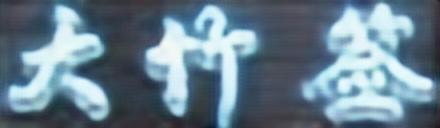}} &
\raisebox{-.5\height}{\includegraphics[height=0.1\textwidth,keepaspectratio]{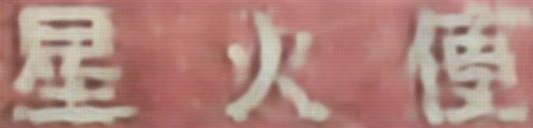}} &
\raisebox{-.5\height}{\includegraphics[height=0.1\textwidth,keepaspectratio]{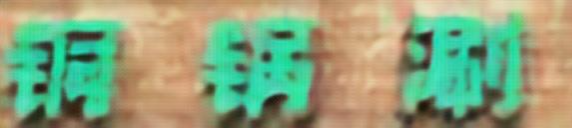}} &
\raisebox{-.5\height}{\includegraphics[height=0.1\textwidth,keepaspectratio]{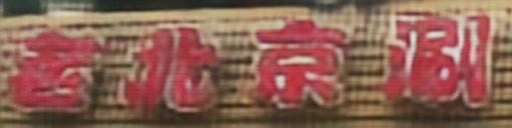}}
\vspace{1mm}
\\

\raisebox{-.5\height}{\makebox[0.2\textwidth][r]{\large TATT}} &
\raisebox{-.5\height}{\includegraphics[height=0.1\textwidth,keepaspectratio]{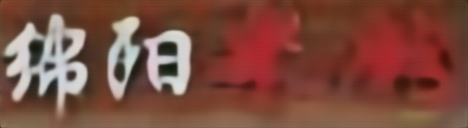}} &
\raisebox{-.5\height}{\includegraphics[height=0.1\textwidth,keepaspectratio]{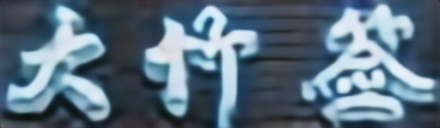}} &
\raisebox{-.5\height}{\includegraphics[height=0.1\textwidth,keepaspectratio]{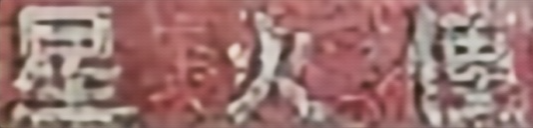}} &
\raisebox{-.5\height}{\includegraphics[height=0.1\textwidth,keepaspectratio]{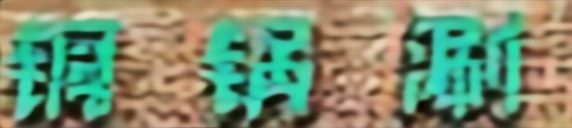}} &
\raisebox{-.5\height}{\includegraphics[height=0.1\textwidth,keepaspectratio]{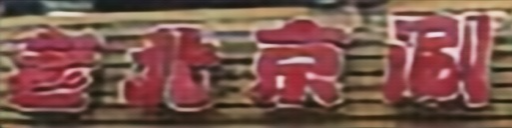}}
\vspace{1mm}
\\

\raisebox{-.5\height}{\makebox[0.2\textwidth][r]{\large StyleSRN}} &
\raisebox{-.5\height}{\includegraphics[height=0.1\textwidth,keepaspectratio]{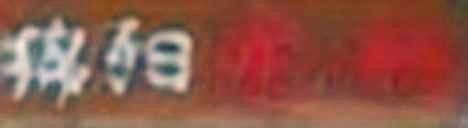}} &
\raisebox{-.5\height}{\includegraphics[height=0.1\textwidth,keepaspectratio]{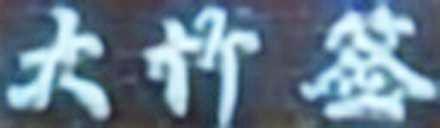}} &
\raisebox{-.5\height}{\includegraphics[height=0.1\textwidth,keepaspectratio]{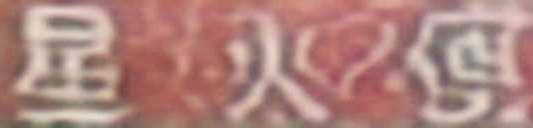}} &
\raisebox{-.5\height}{\includegraphics[height=0.1\textwidth,keepaspectratio]{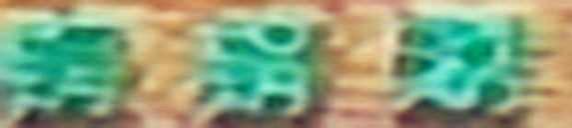}} &
\raisebox{-.5\height}{\includegraphics[height=0.1\textwidth,keepaspectratio]{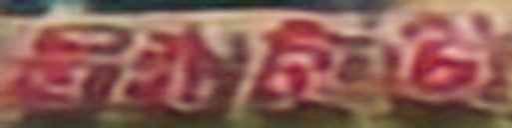}}
\vspace{1mm}
\\

\raisebox{-.5\height}{\makebox[0.2\textwidth][r]{\large MARCONet}} &
\raisebox{-.5\height}{\includegraphics[height=0.1\textwidth,keepaspectratio]{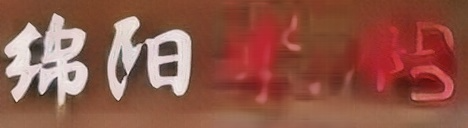}} &
\raisebox{-.5\height}{\includegraphics[height=0.1\textwidth,keepaspectratio]{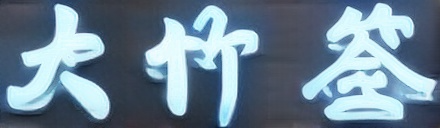}} &
\raisebox{-.5\height}{\includegraphics[height=0.1\textwidth,keepaspectratio]{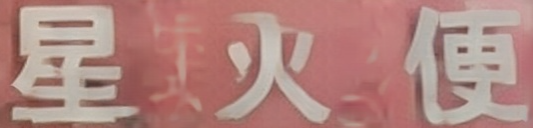}} &
\raisebox{-.5\height}{\includegraphics[height=0.1\textwidth,keepaspectratio]{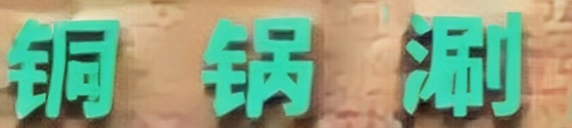}} &
\raisebox{-.5\height}{\includegraphics[height=0.1\textwidth,keepaspectratio]{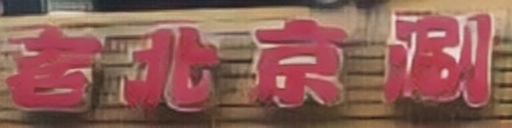}}
\vspace{1mm}
\\

\raisebox{-.5\height}{\makebox[0.2\textwidth][r]{\large DiffTSR}} &
\raisebox{-.5\height}{\includegraphics[height=0.1\textwidth,keepaspectratio]{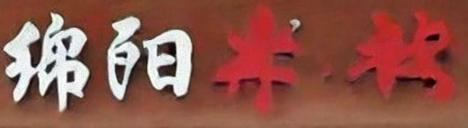}} &
\raisebox{-.5\height}{\includegraphics[height=0.1\textwidth,keepaspectratio]{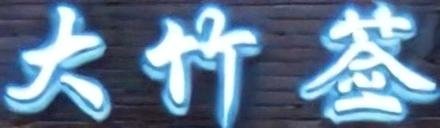}} &
\raisebox{-.5\height}{\includegraphics[height=0.1\textwidth,keepaspectratio]{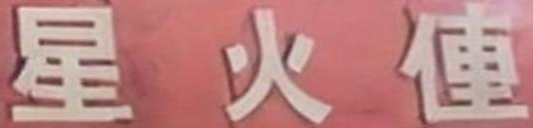}} &
\raisebox{-.5\height}{\includegraphics[height=0.1\textwidth,keepaspectratio]{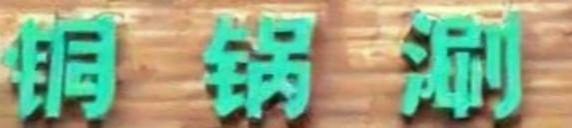}} &
\raisebox{-.5\height}{\includegraphics[height=0.1\textwidth,keepaspectratio]{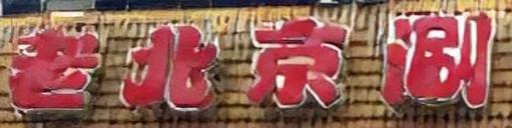}}
\vspace{1mm}
\\

\raisebox{-.5\height}{\makebox[0.2\textwidth][r]{\large TeReDiff}} &
\raisebox{-.5\height}{\includegraphics[height=0.1\textwidth,keepaspectratio]{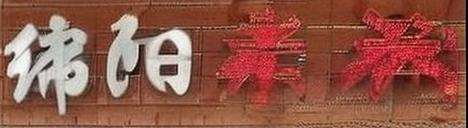}} &
\raisebox{-.5\height}{\includegraphics[height=0.1\textwidth,keepaspectratio]{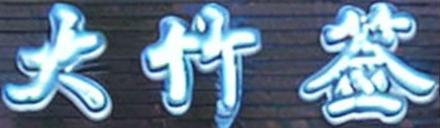}} &
\raisebox{-.5\height}{\includegraphics[height=0.1\textwidth,keepaspectratio]{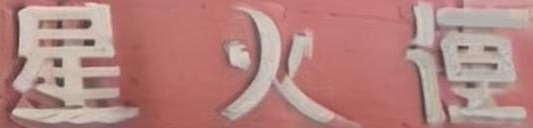}} &
\raisebox{-.5\height}{\includegraphics[height=0.1\textwidth,keepaspectratio]{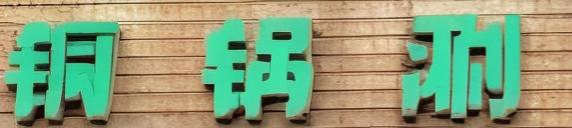}} &
\raisebox{-.5\height}{\includegraphics[height=0.1\textwidth,keepaspectratio]{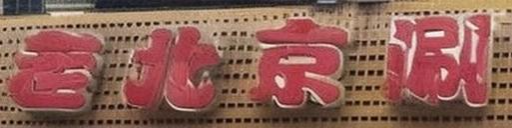}}
\vspace{1mm}
\\

\raisebox{-.5\height}{\makebox[0.2\textwidth][r]{\large PRISM}} &
\raisebox{-.5\height}{\includegraphics[height=0.1\textwidth,keepaspectratio]{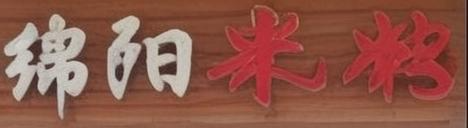}} &
\raisebox{-.5\height}{\includegraphics[height=0.1\textwidth,keepaspectratio]{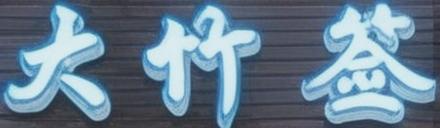}} &
\raisebox{-.5\height}{\includegraphics[height=0.1\textwidth,keepaspectratio]{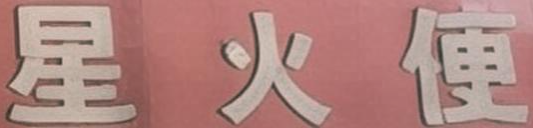}} &
\raisebox{-.5\height}{\includegraphics[height=0.1\textwidth,keepaspectratio]{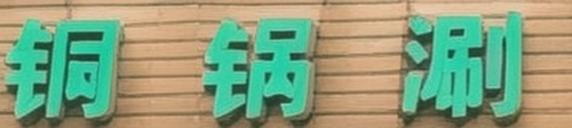}} &
\raisebox{-.5\height}{\includegraphics[height=0.1\textwidth,keepaspectratio]{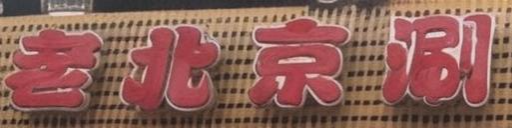}}

\end{tabular}
\end{adjustbox}

\vspace{5mm}
\hspace{-15mm}%
\begin{adjustbox}{max width=1.1\textwidth}
\begin{tabular}{@{}r@{\hspace{2mm}}c@{\hspace{1mm}}c@{\hspace{1mm}}c@{\hspace{1mm}}c@{\hspace{1mm}}c@{\hspace{1mm}}c@{}}

\raisebox{-.5\height}{\makebox[0.2\textwidth][r]{\large GT}} &
\raisebox{-.5\height}{\includegraphics[height=0.1\textwidth,keepaspectratio]{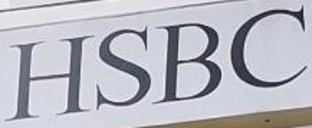}} &
\raisebox{-.5\height}{\includegraphics[height=0.1\textwidth,keepaspectratio]{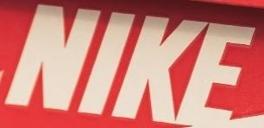}} &
\raisebox{-.5\height}{\includegraphics[height=0.1\textwidth,keepaspectratio]{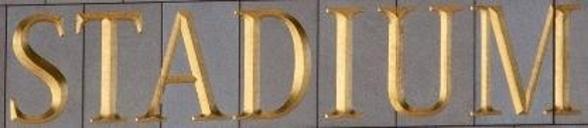}} &
\raisebox{-.5\height}{\includegraphics[height=0.1\textwidth,keepaspectratio]{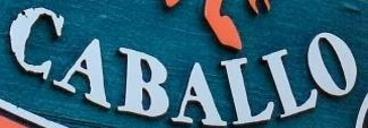}} &
\raisebox{-.5\height}{\includegraphics[height=0.1\textwidth,keepaspectratio]{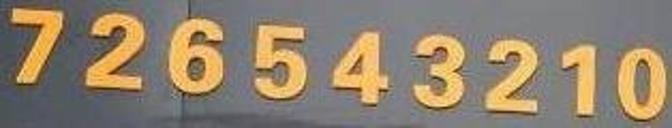}} &
\raisebox{-.5\height}{\includegraphics[height=0.1\textwidth,keepaspectratio]{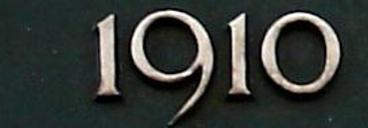}}
\vspace{1mm}
\\

\raisebox{-.5\height}{\makebox[0.2\textwidth][r]{\large LR}} &
\raisebox{-.5\height}{\includegraphics[height=0.1\textwidth,keepaspectratio]{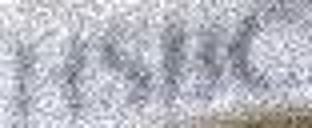}} &
\raisebox{-.5\height}{\includegraphics[height=0.1\textwidth,keepaspectratio]{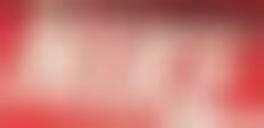}} &
\raisebox{-.5\height}{\includegraphics[height=0.1\textwidth,keepaspectratio]{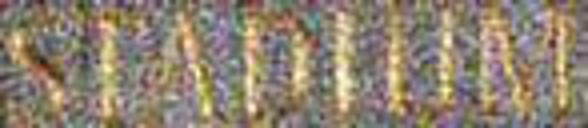}} &
\raisebox{-.5\height}{\includegraphics[height=0.1\textwidth,keepaspectratio]{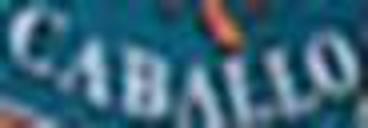}} &
\raisebox{-.5\height}{\includegraphics[height=0.1\textwidth,keepaspectratio]{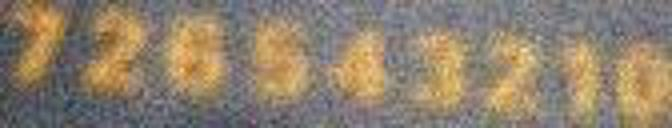}} &
\raisebox{-.5\height}{\includegraphics[height=0.1\textwidth,keepaspectratio]{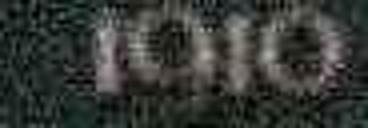}}
\vspace{1mm}
\\

\raisebox{-.5\height}{\makebox[0.2\textwidth][r]{\large TSRN}} &
\raisebox{-.5\height}{\includegraphics[height=0.1\textwidth,keepaspectratio]{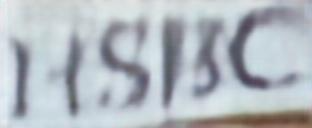}} &
\raisebox{-.5\height}{\includegraphics[height=0.1\textwidth,keepaspectratio]{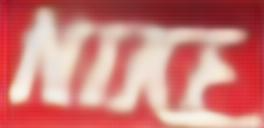}} &
\raisebox{-.5\height}{\includegraphics[height=0.1\textwidth,keepaspectratio]{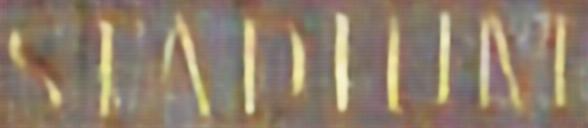}} &
\raisebox{-.5\height}{\includegraphics[height=0.1\textwidth,keepaspectratio]{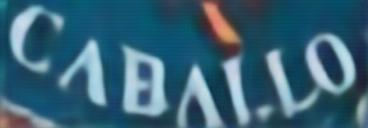}} &
\raisebox{-.5\height}{\includegraphics[height=0.1\textwidth,keepaspectratio]{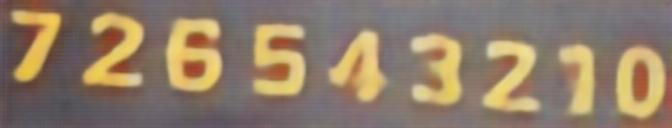}} &
\raisebox{-.5\height}{\includegraphics[height=0.1\textwidth,keepaspectratio]{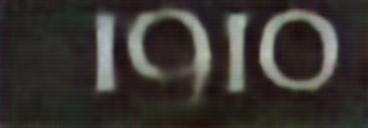}}
\vspace{1mm}
\\

\raisebox{-.5\height}{\makebox[0.2\textwidth][r]{\large TBSRN}} &
\raisebox{-.5\height}{\includegraphics[height=0.1\textwidth,keepaspectratio]{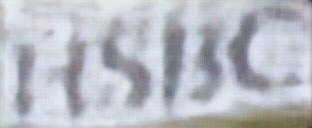}} &
\raisebox{-.5\height}{\includegraphics[height=0.1\textwidth,keepaspectratio]{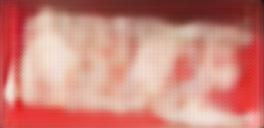}} &
\raisebox{-.5\height}{\includegraphics[height=0.1\textwidth,keepaspectratio]{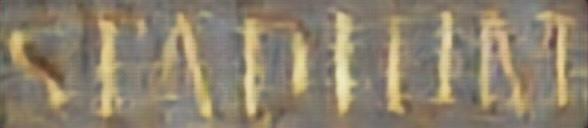}} &
\raisebox{-.5\height}{\includegraphics[height=0.1\textwidth,keepaspectratio]{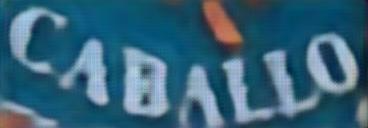}} &
\raisebox{-.5\height}{\includegraphics[height=0.1\textwidth,keepaspectratio]{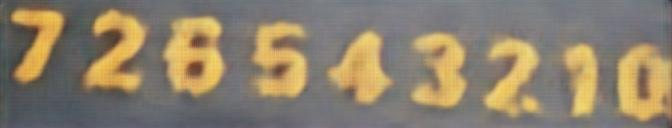}} &
\raisebox{-.5\height}{\includegraphics[height=0.1\textwidth,keepaspectratio]{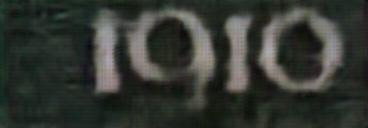}}
\vspace{1mm}
\\

\raisebox{-.5\height}{\makebox[0.2\textwidth][r]{\large TATT}} &
\raisebox{-.5\height}{\includegraphics[height=0.1\textwidth,keepaspectratio]{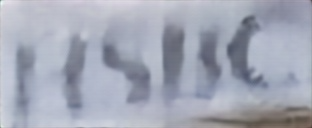}} &
\raisebox{-.5\height}{\includegraphics[height=0.1\textwidth,keepaspectratio]{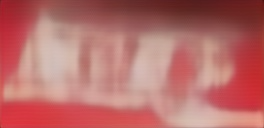}} &
\raisebox{-.5\height}{\includegraphics[height=0.1\textwidth,keepaspectratio]{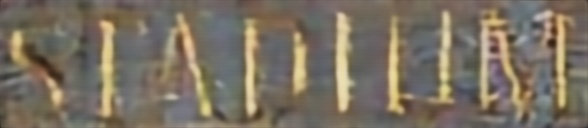}} &
\raisebox{-.5\height}{\includegraphics[height=0.1\textwidth,keepaspectratio]{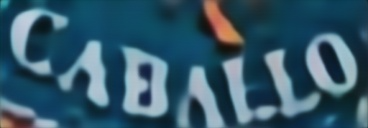}} &
\raisebox{-.5\height}{\includegraphics[height=0.1\textwidth,keepaspectratio]{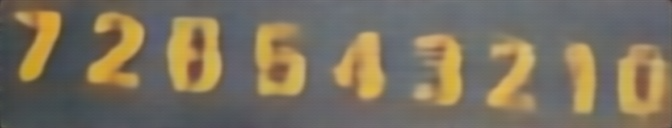}} &
\raisebox{-.5\height}{\includegraphics[height=0.1\textwidth,keepaspectratio]{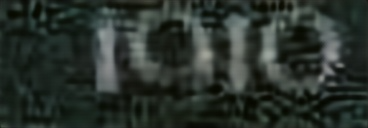}}
\vspace{1mm}
\\

\raisebox{-.5\height}{\makebox[0.2\textwidth][r]{\large StyleSRN}} &
\raisebox{-.5\height}{\includegraphics[height=0.1\textwidth,keepaspectratio]{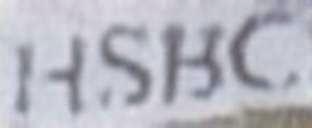}} &
\raisebox{-.5\height}{\includegraphics[height=0.1\textwidth,keepaspectratio]{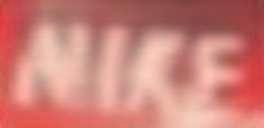}} &
\raisebox{-.5\height}{\includegraphics[height=0.1\textwidth,keepaspectratio]{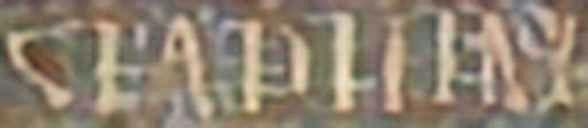}} &
\raisebox{-.5\height}{\includegraphics[height=0.1\textwidth,keepaspectratio]{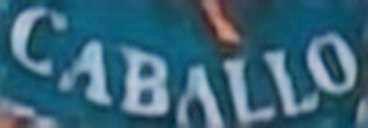}} &
\raisebox{-.5\height}{\includegraphics[height=0.1\textwidth,keepaspectratio]{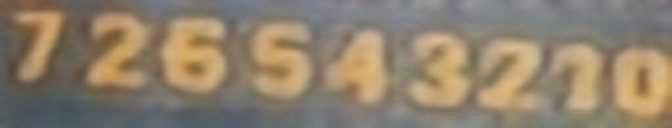}} &
\raisebox{-.5\height}{\includegraphics[height=0.1\textwidth,keepaspectratio]{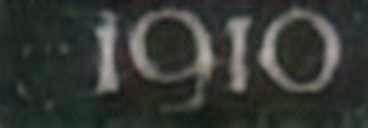}}
\vspace{1mm}
\\

\raisebox{-.5\height}{\makebox[0.2\textwidth][r]{\large MARCONet}} &
\raisebox{-.5\height}{\includegraphics[height=0.1\textwidth,keepaspectratio]{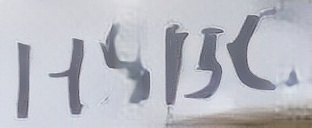}} &
\raisebox{-.5\height}{\includegraphics[height=0.1\textwidth,keepaspectratio]{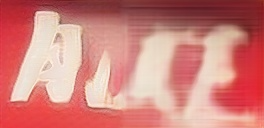}} &
\raisebox{-.5\height}{\includegraphics[height=0.1\textwidth,keepaspectratio]{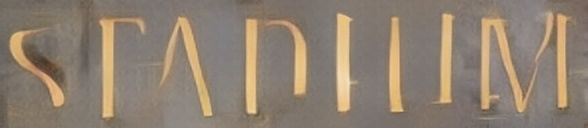}} &
\raisebox{-.5\height}{\includegraphics[height=0.1\textwidth,keepaspectratio]{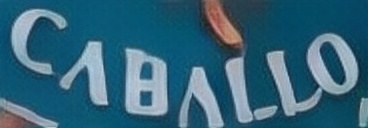}} &
\raisebox{-.5\height}{\includegraphics[height=0.1\textwidth,keepaspectratio]{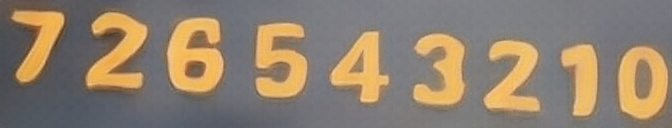}} &
\raisebox{-.5\height}{\includegraphics[height=0.1\textwidth,keepaspectratio]{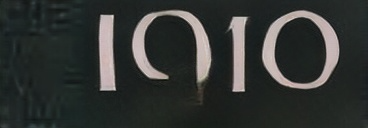}}
\vspace{1mm}
\\

\raisebox{-.5\height}{\makebox[0.2\textwidth][r]{\large DiffTSR}} &
\raisebox{-.5\height}{\includegraphics[height=0.1\textwidth,keepaspectratio]{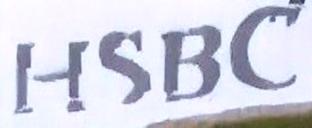}} &
\raisebox{-.5\height}{\includegraphics[height=0.1\textwidth,keepaspectratio]{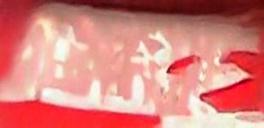}} &
\raisebox{-.5\height}{\includegraphics[height=0.1\textwidth,keepaspectratio]{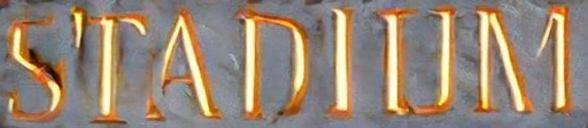}} &
\raisebox{-.5\height}{\includegraphics[height=0.1\textwidth,keepaspectratio]{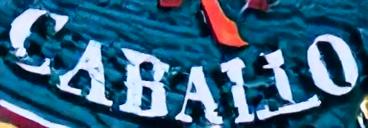}} &
\raisebox{-.5\height}{\includegraphics[height=0.1\textwidth,keepaspectratio]{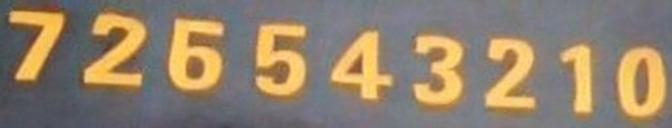}} &
\raisebox{-.5\height}{\includegraphics[height=0.1\textwidth,keepaspectratio]{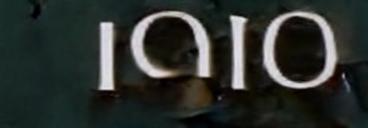}}
\vspace{1mm}
\\

\raisebox{-.5\height}{\makebox[0.2\textwidth][r]{\large TeReDiff}} &
\raisebox{-.5\height}{\includegraphics[height=0.1\textwidth,keepaspectratio]{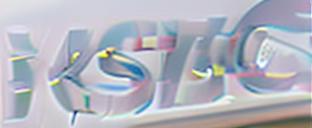}} &
\raisebox{-.5\height}{\includegraphics[height=0.1\textwidth,keepaspectratio]{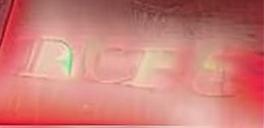}} &
\raisebox{-.5\height}{\includegraphics[height=0.1\textwidth,keepaspectratio]{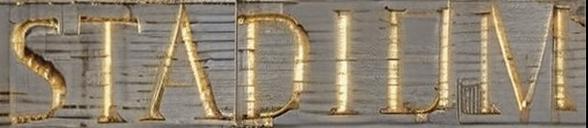}} &
\raisebox{-.5\height}{\includegraphics[height=0.1\textwidth,keepaspectratio]{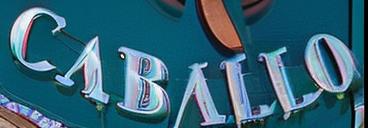}} &
\raisebox{-.5\height}{\includegraphics[height=0.1\textwidth,keepaspectratio]{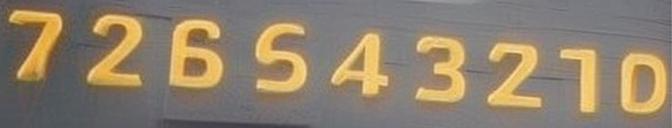}} &
\raisebox{-.5\height}{\includegraphics[height=0.1\textwidth,keepaspectratio]{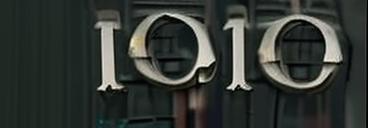}}
\vspace{1mm}
\\

\raisebox{-.5\height}{\makebox[0.2\textwidth][r]{\large PRISM}} &
\raisebox{-.5\height}{\includegraphics[height=0.1\textwidth,keepaspectratio]{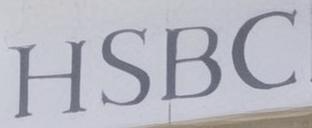}} &
\raisebox{-.5\height}{\includegraphics[height=0.1\textwidth,keepaspectratio]{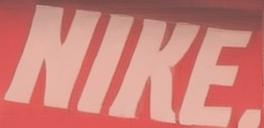}} &
\raisebox{-.5\height}{\includegraphics[height=0.1\textwidth,keepaspectratio]{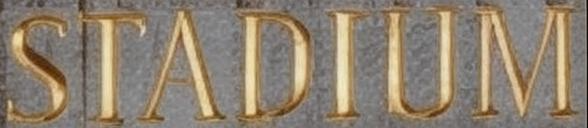}} &
\raisebox{-.5\height}{\includegraphics[height=0.1\textwidth,keepaspectratio]{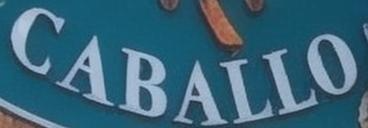}} &
\raisebox{-.5\height}{\includegraphics[height=0.1\textwidth,keepaspectratio]{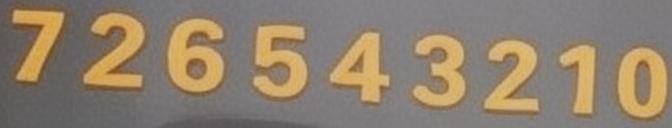}} &
\raisebox{-.5\height}{\includegraphics[height=0.1\textwidth,keepaspectratio]{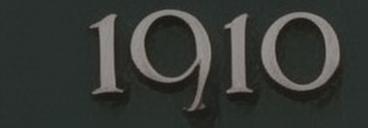}}

\end{tabular}
\end{adjustbox}

\caption{More visualizations on the synthetic dataset BTL-test for $\times 4$ super-resolution.}
\label{fig:app_btl}
\end{figure*}

\begin{figure*}[t]
\centering

\setlength{\tabcolsep}{0pt}
\renewcommand{\arraystretch}{0}

\hspace{-15mm}%
\begin{adjustbox}{max width=1.1\textwidth}
\begin{tabular}{@{}r@{\hspace{2mm}}c@{\hspace{1mm}}c@{\hspace{1mm}}c@{\hspace{1mm}}c@{\hspace{1mm}}c@{}}

\raisebox{-.5\height}{\makebox[0.2\textwidth][r]{\large GT}} &
\raisebox{-.5\height}{\includegraphics[height=0.1\textwidth,keepaspectratio]{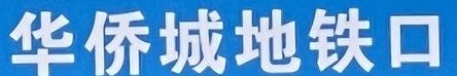}} &
\raisebox{-.5\height}{\includegraphics[height=0.1\textwidth,keepaspectratio]{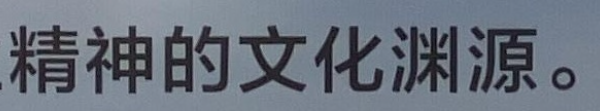}} &
\raisebox{-.5\height}{\includegraphics[height=0.1\textwidth,keepaspectratio]{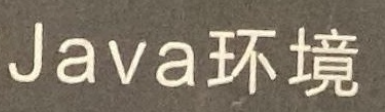}} &
\raisebox{-.5\height}{\includegraphics[height=0.1\textwidth,keepaspectratio]{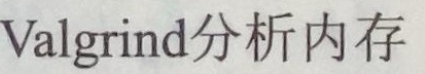}} &
\raisebox{-.5\height}{\includegraphics[height=0.1\textwidth,keepaspectratio]{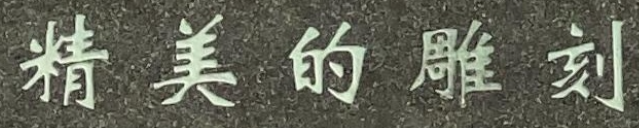}}
\vspace{1mm}
\\

\raisebox{-.5\height}{\makebox[0.2\textwidth][r]{\large LR}} &
\raisebox{-.5\height}{\includegraphics[height=0.1\textwidth,keepaspectratio]{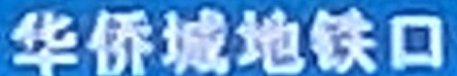}} &
\raisebox{-.5\height}{\includegraphics[height=0.1\textwidth,keepaspectratio]{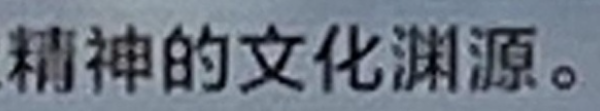}} &
\raisebox{-.5\height}{\includegraphics[height=0.1\textwidth,keepaspectratio]{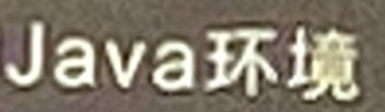}} &
\raisebox{-.5\height}{\includegraphics[height=0.1\textwidth,keepaspectratio]{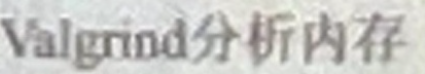}} &
\raisebox{-.5\height}{\includegraphics[height=0.1\textwidth,keepaspectratio]{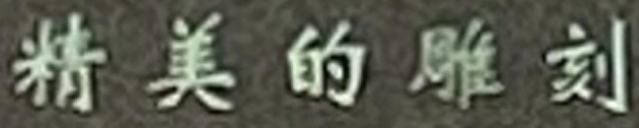}}
\vspace{1mm}
\\

\raisebox{-.5\height}{\makebox[0.2\textwidth][r]{\large TSRN}} &
\raisebox{-.5\height}{\includegraphics[height=0.1\textwidth,keepaspectratio]{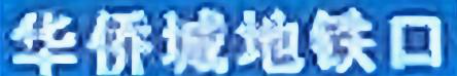}} &
\raisebox{-.5\height}{\includegraphics[height=0.1\textwidth,keepaspectratio]{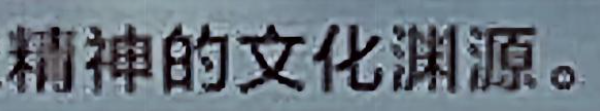}} &
\raisebox{-.5\height}{\includegraphics[height=0.1\textwidth,keepaspectratio]{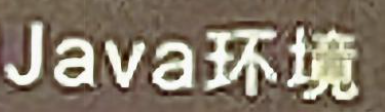}} &
\raisebox{-.5\height}{\includegraphics[height=0.1\textwidth,keepaspectratio]{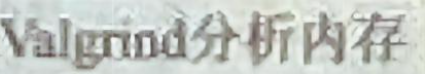}} &
\raisebox{-.5\height}{\includegraphics[height=0.1\textwidth,keepaspectratio]{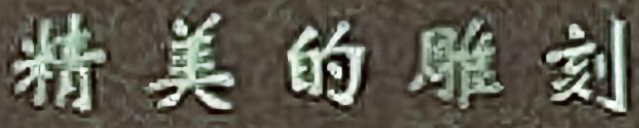}}
\vspace{1mm}
\\

\raisebox{-.5\height}{\makebox[0.2\textwidth][r]{\large TBSRN}} &
\raisebox{-.5\height}{\includegraphics[height=0.1\textwidth,keepaspectratio]{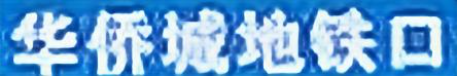}} &
\raisebox{-.5\height}{\includegraphics[height=0.1\textwidth,keepaspectratio]{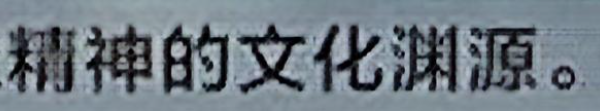}} &
\raisebox{-.5\height}{\includegraphics[height=0.1\textwidth,keepaspectratio]{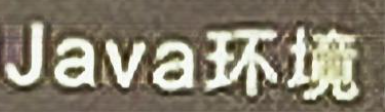}} &
\raisebox{-.5\height}{\includegraphics[height=0.1\textwidth,keepaspectratio]{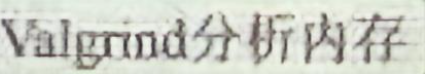}} &
\raisebox{-.5\height}{\includegraphics[height=0.1\textwidth,keepaspectratio]{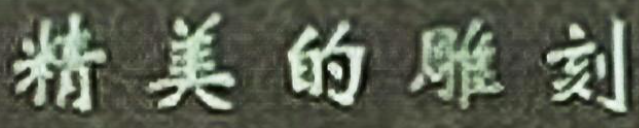}}
\vspace{1mm}
\\

\raisebox{-.5\height}{\makebox[0.2\textwidth][r]{\large TATT}} &
\raisebox{-.5\height}{\includegraphics[height=0.1\textwidth,keepaspectratio]{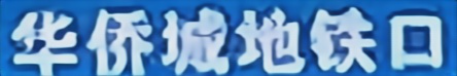}} &
\raisebox{-.5\height}{\includegraphics[height=0.1\textwidth,keepaspectratio]{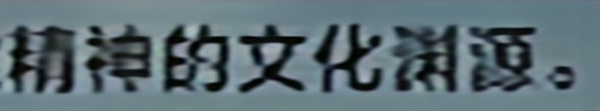}} &
\raisebox{-.5\height}{\includegraphics[height=0.1\textwidth,keepaspectratio]{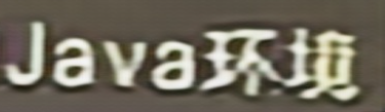}} &
\raisebox{-.5\height}{\includegraphics[height=0.1\textwidth,keepaspectratio]{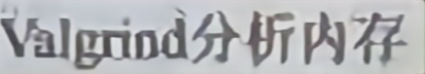}} &
\raisebox{-.5\height}{\includegraphics[height=0.1\textwidth,keepaspectratio]{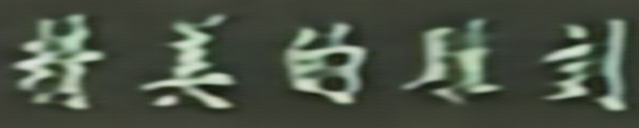}}
\vspace{1mm}
\\

\raisebox{-.5\height}{\makebox[0.2\textwidth][r]{\large StyleSRN}} &
\raisebox{-.5\height}{\includegraphics[height=0.1\textwidth,keepaspectratio]{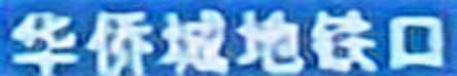}} &
\raisebox{-.5\height}{\includegraphics[height=0.1\textwidth,keepaspectratio]{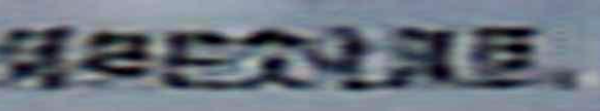}} &
\raisebox{-.5\height}{\includegraphics[height=0.1\textwidth,keepaspectratio]{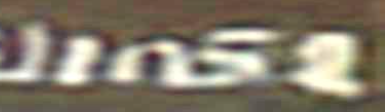}} &
\raisebox{-.5\height}{\includegraphics[height=0.1\textwidth,keepaspectratio]{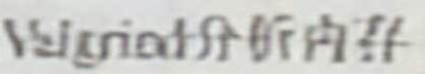}} &
\raisebox{-.5\height}{\includegraphics[height=0.1\textwidth,keepaspectratio]{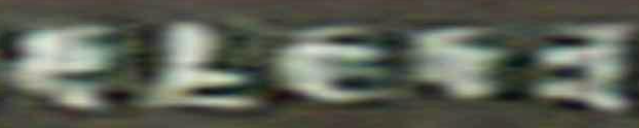}}
\vspace{1mm}
\\

\raisebox{-.5\height}{\makebox[0.2\textwidth][r]{\large MARCONet}} &
\raisebox{-.5\height}{\includegraphics[height=0.1\textwidth,keepaspectratio]{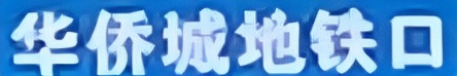}} &
\raisebox{-.5\height}{\includegraphics[height=0.1\textwidth,keepaspectratio]{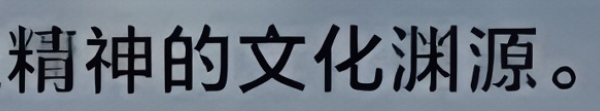}} &
\raisebox{-.5\height}{\includegraphics[height=0.1\textwidth,keepaspectratio]{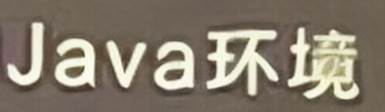}} &
\raisebox{-.5\height}{\includegraphics[height=0.1\textwidth,keepaspectratio]{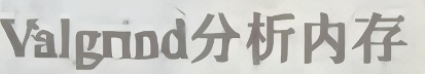}} &
\raisebox{-.5\height}{\includegraphics[height=0.1\textwidth,keepaspectratio]{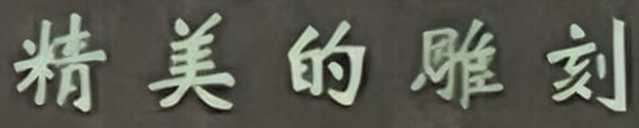}}
\vspace{1mm}
\\

\raisebox{-.5\height}{\makebox[0.2\textwidth][r]{\large DiffTSR}} &
\raisebox{-.5\height}{\includegraphics[height=0.1\textwidth,keepaspectratio]{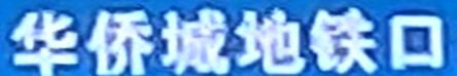}} &
\raisebox{-.5\height}{\includegraphics[height=0.1\textwidth,keepaspectratio]{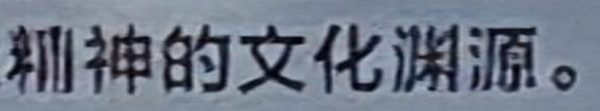}} &
\raisebox{-.5\height}{\includegraphics[height=0.1\textwidth,keepaspectratio]{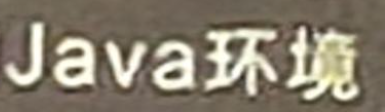}} &
\raisebox{-.5\height}{\includegraphics[height=0.1\textwidth,keepaspectratio]{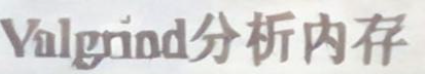}} &
\raisebox{-.5\height}{\includegraphics[height=0.1\textwidth,keepaspectratio]{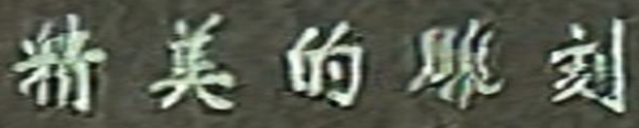}}
\vspace{1mm}
\\

\raisebox{-.5\height}{\makebox[0.2\textwidth][r]{\large TeReDiff}} &
\raisebox{-.5\height}{\includegraphics[height=0.1\textwidth,keepaspectratio]{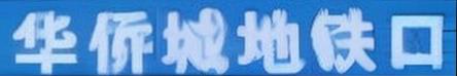}} &
\raisebox{-.5\height}{\includegraphics[height=0.1\textwidth,keepaspectratio]{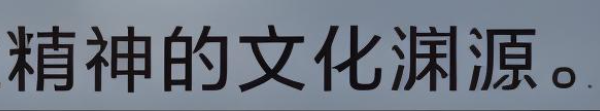}} &
\raisebox{-.5\height}{\includegraphics[height=0.1\textwidth,keepaspectratio]{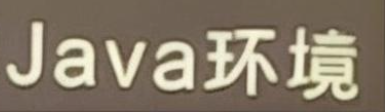}} &
\raisebox{-.5\height}{\includegraphics[height=0.1\textwidth,keepaspectratio]{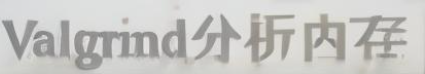}} &
\raisebox{-.5\height}{\includegraphics[height=0.1\textwidth,keepaspectratio]{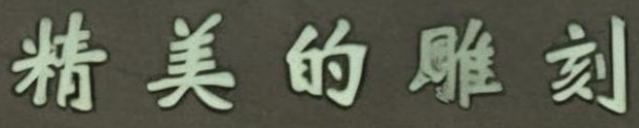}}
\vspace{1mm}
\\

\raisebox{-.5\height}{\makebox[0.2\textwidth][r]{\large PRISM}} &
\raisebox{-.5\height}{\includegraphics[height=0.1\textwidth,keepaspectratio]{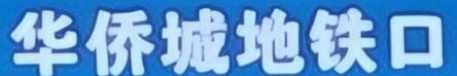}} &
\raisebox{-.5\height}{\includegraphics[height=0.1\textwidth,keepaspectratio]{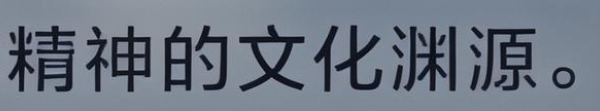}} &
\raisebox{-.5\height}{\includegraphics[height=0.1\textwidth,keepaspectratio]{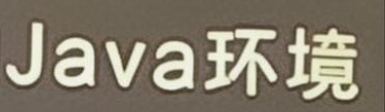}} &
\raisebox{-.5\height}{\includegraphics[height=0.1\textwidth,keepaspectratio]{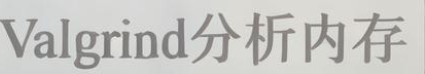}} &
\raisebox{-.5\height}{\includegraphics[height=0.1\textwidth,keepaspectratio]{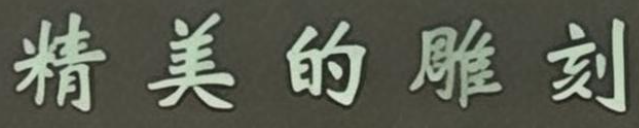}}

\end{tabular}
\end{adjustbox}

\vspace{5mm}

\hspace{-15mm}%
\begin{adjustbox}{max width=1.1\textwidth}
\begin{tabular}{@{}r@{\hspace{3mm}}c@{\hspace{3mm}}c@{\hspace{3mm}}c@{\hspace{3mm}}c@{}}

\raisebox{-.5\height}{\makebox[0.2\textwidth][r]{\large GT}} &
\raisebox{-.5\height}{\includegraphics[height=0.1\textwidth,keepaspectratio]{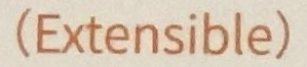}} &
\raisebox{-.5\height}{\includegraphics[height=0.1\textwidth,keepaspectratio]{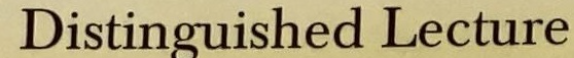}} &
\raisebox{-.5\height}{\includegraphics[height=0.1\textwidth,keepaspectratio]{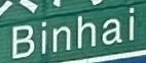}} &
\raisebox{-.5\height}{\includegraphics[height=0.1\textwidth,keepaspectratio]{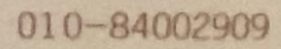}}
\vspace{1mm}
\\

\raisebox{-.5\height}{\makebox[0.2\textwidth][r]{\large LR}} &
\raisebox{-.5\height}{\includegraphics[height=0.1\textwidth,keepaspectratio]{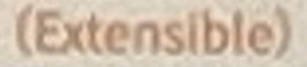}} &
\raisebox{-.5\height}{\includegraphics[height=0.1\textwidth,keepaspectratio]{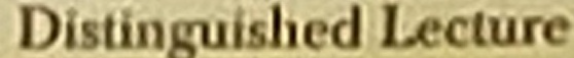}} &
\raisebox{-.5\height}{\includegraphics[height=0.1\textwidth,keepaspectratio]{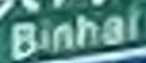}} &
\raisebox{-.5\height}{\includegraphics[height=0.1\textwidth,keepaspectratio]{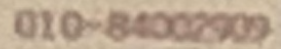}}
\vspace{1mm}
\\

\raisebox{-.5\height}{\makebox[0.2\textwidth][r]{\large TSRN}} &
\raisebox{-.5\height}{\includegraphics[height=0.1\textwidth,keepaspectratio]{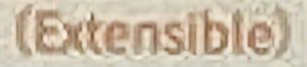}} &
\raisebox{-.5\height}{\includegraphics[height=0.1\textwidth,keepaspectratio]{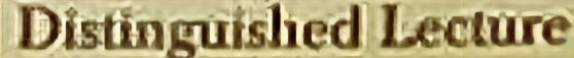}} &
\raisebox{-.5\height}{\includegraphics[height=0.1\textwidth,keepaspectratio]{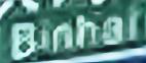}} &
\raisebox{-.5\height}{\includegraphics[height=0.1\textwidth,keepaspectratio]{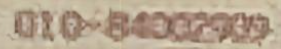}}
\vspace{1mm}
\\

\raisebox{-.5\height}{\makebox[0.2\textwidth][r]{\large TBSRN}} &
\raisebox{-.5\height}{\includegraphics[height=0.1\textwidth,keepaspectratio]{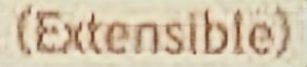}} &
\raisebox{-.5\height}{\includegraphics[height=0.1\textwidth,keepaspectratio]{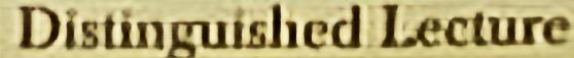}} &
\raisebox{-.5\height}{\includegraphics[height=0.1\textwidth,keepaspectratio]{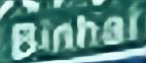}} &
\raisebox{-.5\height}{\includegraphics[height=0.1\textwidth,keepaspectratio]{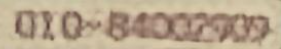}}
\vspace{1mm}
\\

\raisebox{-.5\height}{\makebox[0.2\textwidth][r]{\large TATT}} &
\raisebox{-.5\height}{\includegraphics[height=0.1\textwidth,keepaspectratio]{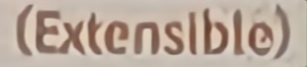}} &
\raisebox{-.5\height}{\includegraphics[height=0.1\textwidth,keepaspectratio]{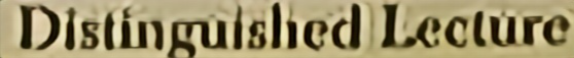}} &
\raisebox{-.5\height}{\includegraphics[height=0.1\textwidth,keepaspectratio]{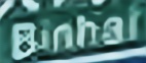}} &
\raisebox{-.5\height}{\includegraphics[height=0.1\textwidth,keepaspectratio]{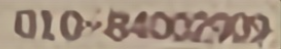}}
\vspace{1mm}
\\

\raisebox{-.5\height}{\makebox[0.2\textwidth][r]{\large StyleSRN}} &
\raisebox{-.5\height}{\includegraphics[height=0.1\textwidth,keepaspectratio]{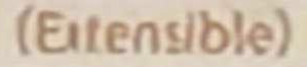}} &
\raisebox{-.5\height}{\includegraphics[height=0.1\textwidth,keepaspectratio]{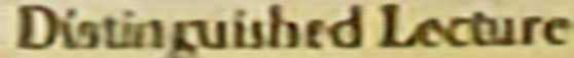}} &
\raisebox{-.5\height}{\includegraphics[height=0.1\textwidth,keepaspectratio]{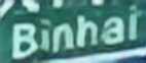}} &
\raisebox{-.5\height}{\includegraphics[height=0.1\textwidth,keepaspectratio]{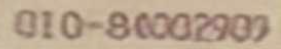}}
\vspace{1mm}
\\

\raisebox{-.5\height}{\makebox[0.2\textwidth][r]{\large MARCONet}} &
\raisebox{-.5\height}{\includegraphics[height=0.1\textwidth,keepaspectratio]{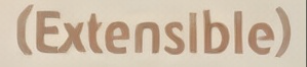}} &
\raisebox{-.5\height}{\includegraphics[height=0.1\textwidth,keepaspectratio]{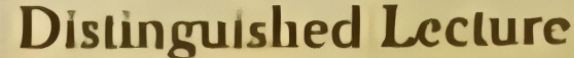}} &
\raisebox{-.5\height}{\includegraphics[height=0.1\textwidth,keepaspectratio]{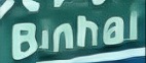}} &
\raisebox{-.5\height}{\includegraphics[height=0.1\textwidth,keepaspectratio]{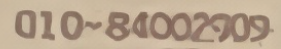}}
\vspace{1mm}
\\

\raisebox{-.5\height}{\makebox[0.2\textwidth][r]{\large DiffTSR}} &
\raisebox{-.5\height}{\includegraphics[height=0.1\textwidth,keepaspectratio]{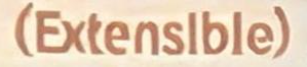}} &
\raisebox{-.5\height}{\includegraphics[height=0.1\textwidth,keepaspectratio]{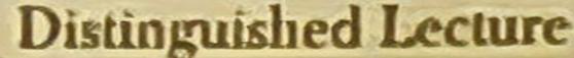}} &
\raisebox{-.5\height}{\includegraphics[height=0.1\textwidth,keepaspectratio]{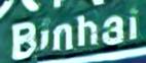}} &
\raisebox{-.5\height}{\includegraphics[height=0.1\textwidth,keepaspectratio]{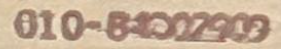}}
\vspace{1mm}
\\

\raisebox{-.5\height}{\makebox[0.2\textwidth][r]{\large TeReDiff}} &
\raisebox{-.5\height}{\includegraphics[height=0.1\textwidth,keepaspectratio]{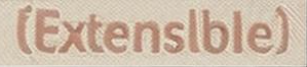}} &
\raisebox{-.5\height}{\includegraphics[height=0.1\textwidth,keepaspectratio]{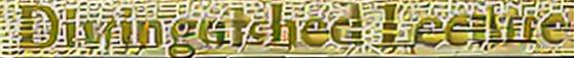}} &
\raisebox{-.5\height}{\includegraphics[height=0.1\textwidth,keepaspectratio]{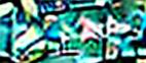}} &
\raisebox{-.5\height}{\includegraphics[height=0.1\textwidth,keepaspectratio]{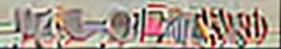}}
\vspace{1mm}
\\

\raisebox{-.5\height}{\makebox[0.2\textwidth][r]{\large PRISM}} &
\raisebox{-.5\height}{\includegraphics[height=0.1\textwidth,keepaspectratio]{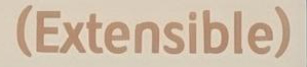}} &
\raisebox{-.5\height}{\includegraphics[height=0.1\textwidth,keepaspectratio]{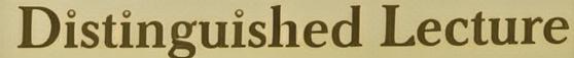}} &
\raisebox{-.5\height}{\includegraphics[height=0.1\textwidth,keepaspectratio]{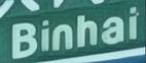}} &
\raisebox{-.5\height}{\includegraphics[height=0.1\textwidth,keepaspectratio]{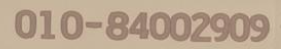}}

\end{tabular}
\end{adjustbox}

\caption{More visualizations on the real-world dataset RealCE-val for $\times 4$ super-resolution.}
\label{fig:app_realce}
\end{figure*}

We provide additional visual comparisons on the synthetic BTL-test set and the real-world RealCE-val set in Figs.~\ref{fig:app_btl} and~\ref{fig:app_realce}. These examples cover diverse text-line types, including Chinese, English, and digit-only samples, as well as both short and long text lines. As shown in the figures, TSRN~\citep{wang2020tsrn}, TBSRN~\citep{chen2021tbsrn}, and TATT~\citep{ma2022tatt} often produce relatively smooth results, especially for structurally complex Chinese characters, where fine strokes and character components are difficult to recover. MARCONet~\citep{li2023marconet} generally improves visual sharpness, but it may introduce distorted glyph shapes, unnatural foreground-background separation, or local stroke artifacts such as broken and merged strokes. DiffTSR~\citep{zhang2024difftsr} can restore plausible text structures in some cases, but under severe degradation it may miss fine strokes or produce incorrect characters. TeReDiff~\citep{min2026terediff} produces sharp outputs, yet it sometimes introduces unrealistic textures, color artifacts, or extra strokes, particularly on small or heavily degraded text images. In contrast, PRISM restores clearer and more coherent text structures across both synthetic and real-world examples. The characters recovered by PRISM show fewer broken or merged strokes, and the text regions are better integrated with the surrounding background. These visual results further indicate that the proposed recoverable prior and uncertainty-aware structure modeling help improve both perceptual clarity and text readability under challenging degradation.

\newpage
\newpage

\vspace{-5mm}%
\section{Architecture Details of PRISM}
\label{app:architecture_details}
\noindent \textbf{Prior Encoder in FMPR.}
The privileged prior encoder $\mathcal{E}_{\mathrm{p}}$ and the LQ-only prior encoder $\mathcal{E}_{\mathrm{lq}}$ share the same architecture except for the input channel number. The input is a latent tensor $z \in \mathbb{R}^{C_{\mathrm{in}} \times \frac{H}{8} \times \frac{W}{8}}$, where $C_{\mathrm{in}}=8$ for $\mathcal{E}_{\mathrm{p}}$ because it takes the channel-wise concatenation of $z_l$ and $z_h$, and $C_{\mathrm{in}}=4$ for $\mathcal{E}_{\mathrm{lq}}$ because it only takes the LQ latent $z_l$. The encoder first maps the input to a 256-channel feature space with a $3{\times}3$ convolution and LeakyReLU, followed by four residual blocks at the same spatial resolution. The feature is then projected to 1024 channels by three $3{\times}3$ convolution layers and adaptively pooled to a fixed spatial size of $4{\times}16$. This yields $N=64$ spatial tokens. After reshaping the feature into a sequence of shape $(B,64,1024)$, a two-layer MLP mixer performs token-wise and channel-wise mixing with LayerNorm. A final linear projection produces the prior embedding in $\mathbb{R}^{B \times 64 \times 1024}$.

\noindent \textbf{Flow-Matching Velocity Network in FMPR.}
The velocity network $\mathcal{V}_{\mathrm{FM}}$ is a lightweight token-wise MLP operating in the prior embedding space. Given the current prior embedding $c^k \in \mathbb{R}^{B \times N \times D}$ and the normalized integration step $k/K$, the scalar timestep is broadcast to $(B,N,1)$ and concatenated with $c^k$ along the feature dimension. A linear layer first maps the resulting $(B,N,D+1)$ representation back to dimension $D=1024$. The representation is then refined by four residual MLP blocks, each consisting of a linear layer and LeakyReLU activation. The network outputs a velocity tensor in $\mathbb{R}^{B \times N \times D}$, which is used in the $K=16$ step Euler integration in Eq.~\eqref{eq:K_step}.

\noindent \textbf{Uncertainty-Aware Spatial Cue Extractor in SURE.}
The spatial cue extractor $\mathcal{F}_{\eta}$ takes the degraded image $x_l \in \mathbb{R}^{B \times 3 \times H \times W}$ as input and produces a projected structural cue $p_s \in \mathbb{R}^{B \times 320 \times \frac{H}{8} \times \frac{W}{8}}$ for the structural residual encoder. It also predicts an auxiliary boundary map $\hat{m} \in \mathbb{R}^{B \times 1 \times H \times W}$ for structure supervision.

The extractor consists of a convolutional stem, five downsampling blocks, and a lightweight Feature Pyramid Network (FPN) for top-down fusion. The stem maps the input image to 32 channels with a $3{\times}3$ convolution, GroupNorm, SiLU, and a residual block. The following downsampling blocks gradually reduce the spatial resolution and produce multi-scale features with channel dimensions $(32,64,128,256,512)$. A lightweight FPN fuses the last three scales in a top-down manner and produces an $\frac{H}{8}\times\frac{W}{8}$ feature map $p_{\mathrm{raw}} \in \mathbb{R}^{B \times 128 \times \frac{H}{8} \times \frac{W}{8}}$. The uncertainty-aware latent head operates on $p_{\mathrm{raw}}$. Two parallel convolutional heads predict the mean $\mu$ and log-variance $\log\sigma^2$ of a latent structural cue distribution. During training, the stochastic structural cue is sampled by the reparameterization in Eq.~\eqref{eq:structure_reparam_compact}. The sampled feature is projected to $p_s$ through a learnable projection layer and sent to the structural residual encoder. In parallel, an edge head decodes the sampled feature into the auxiliary boundary map $\hat{m}$. During inference, we use a noise-attenuated stochastic cue derived from the predicted distribution for stable structure control.

\noindent \textbf{Structural Residual Encoder in SURE.}
The structural residual encoder $\mathcal{C}_{\eta}$ follows a ControlNet-style residual conditioning design and is initialized from the encoder part of the diffusion backbone. It takes the degraded latent $z_l$, the recovered prior $\hat{c}$, and the projected structural cue $p_s$ as inputs, and predicts multi-level residual controls $\mathcal{R}=\{r_i\}_{i=1}^{M}$, where $M$ equals 9 according to the UNet structure. These residuals are injected into the skip-connection features of the frozen UNet $\mathcal{U}_{\bar{\theta}}$. Since both the FMPR pathway and the restoration backbone are frozen in this stage, $\mathcal{C}_{\eta}$ focuses on residual spatial refinement rather than re-estimating the text-aware prior.

\section{Broader Impacts and Limitations}
\label{sec:broader_impacts}
\paragraph{Broader impacts.}
This work aims to improve the readability and visual quality of degraded text images. PRISM may benefit applications such as document enhancement, scene text recognition, assistive reading, and OCR preprocessing. Since text super-resolution may reconstruct plausible content from ambiguous inputs, restored results should be used with caution in sensitive scenarios. For legal, medical, financial, or privacy-related use cases, they should be regarded as auxiliary references rather than authoritative evidence.

\paragraph{Limitations.}
Our study focuses on Chinese-English text-line super-resolution with moderate to long aspect ratios. This setting covers many practical text-image cases, but does not fully include full-scene text restoration, dense multi-line documents, or highly irregular text layouts. Extending PRISM to broader text-aware restoration scenarios is a promising future direction. In addition, although BTL combines curated real crops and rendered text lines, its language coverage is still mainly limited to Chinese, English, and digit-based text. Future work may further expand the dataset to more languages, scripts, fonts, and real-world capture conditions.

\end{document}